%% file: acl_latex.tex
\documentclass[11pt]{article}

\usepackage{acl}

\usepackage{times}
\usepackage{latexsym}

\usepackage[T1]{fontenc}

\usepackage[utf8]{inputenc}

\usepackage{microtype}

\usepackage{inconsolata}

\usepackage{graphicx}

\usepackage{comment}
\usepackage{enumitem}
\usepackage{nicefrac}
\usepackage{xcolor}
\usepackage[most]{tcolorbox}
\usepackage{amsmath}
\usepackage{wrapfig}
\usepackage{multirow}
\usepackage{booktabs, makecell, longtable}
\usepackage{soul}
\usepackage{xcolor}
\usepackage{listings}
\usepackage{hyperref}
\usepackage{cleveref}
\usepackage{tabularx}
\usepackage{xcolor}
\usepackage{bm}      
\usepackage{amssymb} 
\newcommand{\dup}{{\large\textcolor{green!55!black}{$\bm{+}$}}} 
\newcommand{\deq}{{\large\textcolor{gray!50}{$\bm{=}$}}}         
\newcommand{\ddn}{{\large\textcolor{red}{$\bm{-}$}}}            
\input{appendix_prompts_preamble}
\newcommand{\blfootnote}[1]{%
  \begingroup
    \renewcommand\thefootnote{}\footnote{#1}%
    \addtocounter{footnote}{-1}%
  \endgroup
}

\title{Playing with Words, Improving with Rewards:\\ Training Language Models for Creative Association}

\author{
  Vijeta Deshpande\textsuperscript{1,*} \quad
  Namrata Shivagunde\textsuperscript{1} \quad
  Sherin Muckatira\textsuperscript{1} \quad
  Hadrien Glaude\textsuperscript{2} \\
  \textbf{Mikhail Gronas\textsuperscript{3}} \quad
  \textbf{Claire Stevenson\textsuperscript{4}} \quad
  \textbf{Roger Beaty\textsuperscript{5}} \quad
  \textbf{Anna Rumshisky\textsuperscript{1,6}} \\
  \textsuperscript{1}University of Massachusetts Lowell \quad
  \textsuperscript{2}Zaqa.ai \quad
  \textsuperscript{3}Dartmouth College \\
  \textsuperscript{4}University of Amsterdam \quad
  \textsuperscript{5}Pennsylvania State University \quad
  \textsuperscript{6}Amazon AGI
}

\begin{document}
\maketitle
\begin{abstract}
Large Language Models (LLMs) are being applied to increasingly difficult problems and use cases. To navigate their vast solution spaces effectively, LLMs need to be creative. Yet the subjective nature of creativity and the limits of human judgment make training LLMs for creativity especially challenging. As a solution, we train LLMs on Codenames, a word-association game that exercises the two central axes of creativity, divergent and convergent thinking, while yielding objectively verifiable outcomes. This verifiability lets us bypass human judgment and train with Reinforcement Learning with Verifiable Rewards (RLVR). We train Qwen3-1.7B, 4B, and 8B models and evaluate them on ten creativity and four reasoning benchmarks. We find that the precision-diversity trade-off is scale-dependent: the 8B model prioritizes creativity over precision, while the 1.7B and 4B models gain reasoning precision at the cost of creativity. Concretely, the 8B model shows modest but consistent creativity gains (8 of 10 benchmarks) with only minor reasoning degradation, whereas the smaller models achieve substantial gains on reasoning tasks. Our study presents a scalable and effective solution to train LLMs for creativity.

\end{abstract}

\section{Introduction}

Large language models\blfootnote{\raggedright\textsuperscript{*}\,Corresponding author:\ \texttt{vijeta\_deshpande@student.uml.edu}} (LLMs) have advanced at a remarkable pace, expanding from fluent text generation to tackling challenging tasks such as solving complex mathematical problems to generating novel content \citep{bubeck2023paper, ahn2024large, wu2026writingbench}. Performance on these hard problems hinges on creativity, the capacity to navigate vast problem spaces and arrive at unconventional yet effective solutions \citep{yang2022creative, beaty2023associative, tian2024macgyver}. This exposes a central but unresolved question in contemporary AI research: how can we explicitly train LLMs to be genuinely creative, rather than merely hoping creativity emerges as a byproduct of scale?

A natural starting point comes from cognitive science, where human creativity has long been operationalized as an associative process over semantic memory \citep{beaty2023associative, mednick1962associative, luchini2025creativity}. Novel ideas arise when distantly related concepts are combined into new, functional wholes. Computational and neurocognitive studies converge on a striking finding that highly creative individuals possess more densely connected semantic networks with shorter path distances between concepts, enabling the rapid linkage of remote ideas that underlie creative association \citep{kenett2019semantic, kenett2014investigating}. This architecture aligns with the principle of combinatorial creativity, where novelty emerges from the rearrangement and integration of existing knowledge components rather than from creation ex nihilo \citep{boden1998creativity, yang2022creative}. LLMs, despite their architectural differences, also learn predominantly through association \citep{mikolov2013distributed, johnson2025large, barenholtz2026world}. 
This parallel motivates a concrete research question: can we systematically train LLMs to integrate the existing knowledge to improve not just creativity but complex problem-solving more broadly?

The central obstacle to training creativity directly is that creativity is inherently subjective \citep{vessel2018stronger, forthmann2017missing}. Humans do not always agree on what constitutes creative work \citep{vessel2018stronger, cortes2021inconsistency}. We sidestep this difficulty \citep{casper2023open} by turning to a structured environment where creative decision-making is inherent, but the outcomes of the decisions are objectively measurable: the word-association game \href{https://en.wikipedia.org/wiki/Codenames_(board_game)}{Codenames}. Codenames externalizes the two canonical stages of creative cognition — divergence (generating remote associations that span multiple concepts) and convergence (selecting the single best association that uniquely identifies the targets) — and grades each turn with an unambiguous outcome \citep{cropley2006praise}. This makes it an ideal testbed for putting an LLM's associative capacity under measurable pressure without the noise of subjective preference judgments.

In Codenames, a board of unrelated words is laid out, and a small subset is secretly designated as target words; the remaining words on the board, including some that must be actively avoided, are non-targets. One player, the spymaster, sees which words are targets and must produce a single clue word that points to as many targets as possible while steering teammates away from every non-target. The teammates, who see only the board, then guess which words the clue refers to. A good clue is semantically remote enough from the non-targets to disambiguate them, yet semantically close enough to several targets to link them through a shared latent dimension — a chair, an altar, and a stage might all be connected through the abstract concept of platform. Strong play, therefore, demands exactly the divergent-then-convergent associative leap that cognitive scientists identify as the core mechanism of creativity.


Casting creativity as a game with verifiable outcomes lets us bypass expensive human preference annotation entirely. Instead, we train with Reinforcement Learning with Verifiable Rewards (RLVR) \citep{guo2025deepseek}, the paradigm that has recently driven step-change improvements in LLMs. Through RLVR training on Codenames, we fine-tune three Qwen3 models (1.7B, 4B, and 8B) to reorganize existing knowledge and discover unseen associations. We summarize our key contributions and findings below:

\begin{itemize}[leftmargin=10pt, noitemsep, topsep=2pt]
    \item \textbf{Creativity Training.} We cast creativity as a verifiable-reward game, train LLMs via RLVR without human annotation, and evaluate on ten creativity tasks and four reasoning benchmarks.
    \item \textbf{Precision-Diversity Trade-off.} We show this known trade-off is scale-dependent: smaller models gain reasoning precision at the cost of creativity, while the 8B model gains creativity with a slight precision loss.
    \item \textbf{Creativity Gains.} The 8B model shows modest but consistent creativity gains.
    \item \textbf{Reasoning Gains.} The 1.7B and 4B models show substantial gains on reasoning tasks.
\end{itemize}

\section{Related Work}

\paragraph{Improving LLM Creativity.}
Prior work has explored several routes toward more creative LLMs. A first line uses prompting and decoding strategies \cite{tian2024macgyver, mehrotra2024enhancing, nair-etal-2024-creative, summers-stay2023brainstorm, franceschelli2024creative, meister2023locally}, including inference-time analogical reasoning that searches for cross-domain solutions \citep{shen2026unlocking}. A second line uses multi-agent role-playing, such as the three-phase divergent-to-convergent ideation framework of \citet{lu2024llm}. A third line turns to learning: \citet{fukuda-etal-2025-building} fine-tune on the Divergent Association Task and find gains only under specific conditions, while \citet{ismayilzada2025creative} proposes CrPO, a preference-based alignment method optimizing novelty, diversity, and surprise on human creativity judgments. While some studies focus on data selection strategies to improve various dimensions of creativity \citep{lanchantin2025diverse, deshpande2025diverse}. Across these efforts, RLVR \citep{guo2025deepseek}, the method most responsible for recent step-change gains and the most scalable since it needs no human preference annotation, remains unexplored as a route to creativity.

\paragraph{Games and Self-Improvement with RLVR.}
A core limitation of RLHF \citep{ouyang2022training} is its dependence on human-curated data, which constrains scalability~\cite{liu2025spiral, wang2025vision}. Training LLMs on games through self-play with verifiable rewards (RLVR) holds the potential to address this bottleneck to a great extent \cite{schaul2024boundless}. A growing body of work has applied this paradigm across diverse environments, e.g., adversarial word games \cite{cheng2024self}, logic puzzles \citep{xie2025logic, wong2025logicpuzzlerl}, strategic board games \citep{hwang2025can, yuan2025marshal}, arcade-style visual games \citep{xie2025play, tong2025game}, and zero-sum multi-turn games \citep{liu2025spiral}—and a striking commonality is the consistent report of out-of-distribution (OOD) gains on reasoning benchmarks unrelated to the training games. 

\section{Method}

\subsection{Training Data}
\label{sec:data}

\paragraph{Game Environment}
\label{sec:data-mdp}
We adopt Codenames, a word association game, as the training environment. The original board game places four sets of words on the board and involves two competing teams. Of these four sets, two are target sets, one assigned to each team. The remaining words form a neutral set and a single assassin word that ends the game. We simplify this setup by removing the neutral and assassin words. This allows us to focus on a clean signal of word association, free from higher-level rules. Each team consists of one clue generator (the \emph{Spymaster}) and one or more guess generators (the \emph{Operatives}). Only the clue generator can see which target words belong to which team. Its goal is to find a common link, the clue, that connects multiple words in its team's target set. The guess generators, by contrast, cannot see how the separation between the two target sets. They have access to all words on the board along with the clue, and their goal is to deduce which words the clue refers to. Each team aims to correctly guess all of its target words before the opposing team. 

In our training setup, we train a single LLM that plays the role of one team in the game environment. This is equivalent to having the LLM switch between both teams, intending to complete its assigned target set as quickly as possible while avoiding the words in the non-target set. In its natural form, Codenames for a single team is a multi-turn discussion. The clue generator has access to all context, while the guess generators cannot see the separation between the two target sets. These discussions can become extremely long. Thinking traces alone can easily reach several thousand tokens for a single turn. This increased length raises training cost and makes training instabilities more likely. To address this, we break the multi-turn discussions into multiple single-turn conversations. Specifically, we model clue generation and guess generation as single-turn tasks, and we strategically sample board positions to simulate every stage of the full end-to-end game. 

More precisely, we model this training environment as a Markov Decision Process (MDP). Let $w$ be any whitespace-separated English word that can be represented by one or more tokens in the LLM's vocabulary $\mathcal{V}$. Let $\mathcal{T}$ and $\mathcal{N}$ denote the target and non-target word sets, and let $\mathcal{B}$ denote the combined and shuffled set $\mathcal{T} \cup \mathcal{N}$. Let $w_{clue}$ be a single whitespace-separable clue word, let $\mathcal{G}$ be the set of guesses, and $g_{max}$ be the maximum number of allowed guesses in one chance. We represent the state set of the MDP as
\begin{equation}
    \label{eq:state_set}
    \mathcal{S} = \{\mathcal{T}, \mathcal{N}, \mathcal{B}, w_{clue}, g_{max}\}.
\end{equation}
During the clue generation task, $\mathcal{B}$, $w_{clue}$, $g_{max}$ are null. During the guess generation task, $\mathcal{T}$ and $\mathcal{N}$ are empty. We define the action set as
\begin{equation}
    \label{eq:action_set}
    \mathcal{A} = \{w_{clue}, \mathcal{T}_{selected}, \mathcal{G}\}.
\end{equation}
During the clue generation task, the LLM generates $w_{clue}$ for a subset of words in $\mathcal{T}$ i.e., $\mathcal{T}_{selected}$ (so $\mathcal{G} = \emptyset$). During the guess generation task, it generates only $\mathcal{G}$ (so $w_{clue} = \mathcal{T}_{selected} = \emptyset$). Note that, $\mathcal{T}_{selected} \subset \mathcal{T}$ and $1 \leq |\mathcal{T}_{selected}|=g_{max}$. We assume that the transition function $\mathcal{P}$ is not accessible for this MDP. In the following subsections, we present details on how we sample our state set and subsequently discuss details on reward functions. 

\paragraph{Sampling State Set}
\label{sec:data-board-positions}
We loosely refer to both $\mathcal{B}$ and $\{\mathcal{T}, \mathcal{N}\}$ as board positions. The two differ in one important way: $\{\mathcal{T}, \mathcal{N}\}$ encodes the separation between $\mathcal{T}$ and $\mathcal{N}$, whereas $\mathcal{B}$ does not. Accordingly, $\{\mathcal{T}, \mathcal{N}\}$ is available only during clue generation, and $\mathcal{B}$ only during guess generation. A complete state also includes the clue word $w_{clue}$, which is required during guess generation. In a natural game, this clue would come from a preceding clue-generation turn. Because we train each turn as an independent single-turn task, no such preceding turn exists, so we must supply $w_{clue}$ in advance. We therefore generate clue words automatically through a systematic three-step process.

\begin{itemize}[leftmargin=10pt, noitemsep, topsep=2pt]
    \item \emph{Generate a Pool of Diverse Topics}: We begin by manually creating a seed list of $50$ diverse topics, such as ocean, food, hobbies etc. With the help of Claude Opus, we then extend this list to roughly $1{,}000$ unique topics. Each topic is a single whitespace-separable word that later serves as the clue in guess generation training examples. 
    \item \emph{Find Words Associated with Topics}: For each topic, we use Claude Opus to generate a list of directly associated words, which we stratify by frequency into common and rare words. For example, the topic \emph{ocean} yields common words such as \emph{tide}, \emph{wave}, and \emph{salt}, and rare words such as \emph{pelagic}, \emph{benthic}, and \emph{halocline}. We store the top $10$ common and $10$ rare words per topic in a repository $\mathcal{R}$. For any topic $t$, $\mathcal{R}[t] = [W^{(t, c)}, W^{(t, r)}]$, where $W^{(t, c)}$ and $W^{(t, r)}$ are the common and rare words associated with $t$, and $|W^{(t, c)}| = |W^{(t, r)}| = 10$.
    \item \emph{Sample State Sets}: To create a board position, we sample two topics, one for the words in $\mathcal{T}$ and one for the words in $\mathcal{N}$. This lets us maintain a coherent theme and a clear separation between the words on the board, which keeps the task tractable for LLMs, especially smaller ones. We then draw the size of each set independently from $[2, 6]$ (inclusive) and sample that many words from the corresponding topic. Varying the set sizes lets us simulate the early and late game states encountered in a long single-turn progression. Finally, the topic word for $\mathcal{T}$ serves as the clue for the words in $\mathcal{T}$. Following this process, we sample the state sets (see \Cref{eq:state_set}) used to train the LLMs.
\end{itemize}

\paragraph{Difficulty of States}
\label{sec:data-difficulty}
When sampling the state sets, we apply four rules of increasing difficulty. Each rule changes only how the words in $\mathcal{T}$ are sampled, while the rule for $\mathcal{N}$ stays fixed. Let $t_{\mathcal{T}}$ and $t_{\mathcal{N}}$ denote the topics sampled for the target and non-target sets, and let $n_{\mathcal{T}} \sim U(2, 6)$ and $n_{\mathcal{N}} \sim U(2, 6)$ denote the number of words sampled for each set. The $n_{\mathcal{N}}$ non-target words are always drawn from the common pool $W^{(t_{\mathcal{N}}, c)}$. The four rules control the target set as follows.

\begin{itemize}
    \item \emph{Simple}: all $n_{\mathcal{T}}$ words are drawn from the common pool $W^{(t_{\mathcal{T}}, c)}$.
    \item \emph{Moderate}: $\lfloor n_{\mathcal{T}}/2 \rfloor$ words are drawn from the common pool $W^{(t_{\mathcal{T}}, c)}$, and the rest from the rare pool $W^{(t_{\mathcal{T}}, r)}$.
    \item \emph{Advanced}: all $n_{\mathcal{T}}$ words are drawn from the rare pool $W^{(t_{\mathcal{T}}, r)}$.
    \item \emph{Expert}: $\lfloor n_{\mathcal{T}}/2 \rfloor$ words are drawn from the rare pool $W^{(t_{\mathcal{T}}, r)}$, and the rest from the common pool of the non-target topic, $W^{(t_{\mathcal{N}}, c)}$.
\end{itemize}
We order the training data from the \emph{Simple} rule to the \emph{Expert} rule, forming a natural curriculum for the LLM. Intuitively, difficulty rises as target words shift from common to rare, and the \emph{Expert} rule adds distractors that overlap with the non-target topic. Moreover, in all but the last rule, $\mathcal{T}_{selected}=\mathcal{T}$ and $g_{max} = |\mathcal{T}|$. In the case of the last rule, $\mathcal{T}_{selected} \subset \mathcal{T}$ and $g_{max} = \lfloor n_{\mathcal{T}}/2 \rfloor$.

\paragraph{Data Statistics}
Overall, we create $500$ unique states per rule for each task, clue generation and guess generation. In total, this yields $4{,}000$ examples, of which we use $3{,}680$ for training and $320$ for validation, balanced across rules and tasks. We represent our training dataset as a collection of $N = 3680$ examples, \begin{equation} \mathcal{D} = \left\{ \left( \mathcal{T}, \mathcal{N}, w_{clue}, g_{\max}, \mathcal{T}_{\mathrm{sel}} \right)_i \right\}_{i=1}^{N}, \end{equation} where each tuple stores the target set $\mathcal{T}$, the non-target set $\mathcal{N}$, the precomputed clue $w_{clue}$, the maximum number of guesses $g_{\max}$, and the selected target subset $\mathcal{T}_{\mathrm{sel}} \subseteq \mathcal{T}$ that the clue is intended for. For exact prompts for each task, see \Cref{app:sec:prompts}.

\subsection{Reward Function}
\label{sec:reward-fn}

\paragraph{Clue Generation Task Reward} Let $\pi$ be the actor policy we train, which maps a state to an action over the spaces defined in \Cref{eq:state_set} and \ref{eq:action_set} ($\pi : \mathcal{S} \rightarrow \mathcal{A}$). To reward a generated clue, we run the full clue-then-guess loop and score the resulting guesses, which proceeds in three stages. First, in the clue generation state $\mathcal{S} = \{\mathcal{T}, \mathcal{N}\}$, the actor samples a clue for $\mathcal{T}_{selected}$ words conditioned on the separated target and non-target sets, ${ w_{clue}, \mathcal{T}_{selected} } \sim \pi( \mathcal{T}, \mathcal{N})$. Second, we form the shuffled board $\mathcal{B} = \mathcal{T} \cup \mathcal{N}$ from the same two sets, pair it with the clue, and a bound for guesses $g_{max} = |\mathcal{T}_{selected}|$ to obtain the guess generation state $\mathcal{S} = \{\mathcal{B}, w_{clue}, g_{max}\}$, and pass this to a frozen judge policy $\phi$ that returns a set of guesses, $\mathcal{G} \sim \phi(\mathcal{B}, w_{clue}, g_{max})$. Third, we compare the guesses against the ground-truth sets $\mathcal{T}$ and $\mathcal{N}$ to compute a scalar reward $r = f(\mathcal{T}, \mathcal{N}, \mathcal{G})$.

The reward credits guesses that hit the target set and penalizes guesses that hit the non-target set or fall outside the board entirely:
\begin{equation}
    \label{eq:reward-fn}
    f(\mathcal{T}, \mathcal{N}, \mathcal{G}) = \frac{|\mathcal{T} \cap \mathcal{G}|}{|\mathcal{T}|} - \frac{|\mathcal{N} \cap \mathcal{G}|}{|\mathcal{N}|} - \frac{|\mathcal{G} \setminus (\mathcal{T} \cup \mathcal{N})|}{|\mathcal{G}|}
\end{equation}
The first term is the fraction of target words correctly recovered, the second is the fraction of non-target words mistakenly selected, and the third is the fraction of guesses that appear in neither set. The reward is therefore bounded in $[-2, 1]$, attaining its maximum of $1$ only when the judge recovers every target word and selects nothing else.

\paragraph{Guess Generation Task Reward} For guess generation, the actor policy produces guesses conditioned on the board and an a priori clue stored in our dataset, $\mathcal{G} \sim \pi( \mathcal{B}, w_{clue})$. For each guess generation example, we also retain the sets $\mathcal{T}$ and $\mathcal{N}$ that compose $\mathcal{B}$. After generation, we therefore have $\mathcal{T}$, $\mathcal{N}$, and $\mathcal{G}$, which is everything needed to compute the reward using the same function as before (\Cref{eq:reward-fn}).

\paragraph{Reward for Response Format} Before computing the task reward, we run format checks on each response. For instance, a clue must not appear on the board. We apply $6$ such checks to clues and $4$ to guesses (see \Cref{app:sec:format-reward}). The task reward (\Cref{eq:reward-fn}) is computed only when all task-specific checks pass; otherwise, the reward is $-1$.

\paragraph{Summary} In summary, we cast our problem as a finite-horizon MDP, \begin{equation} \mathcal{M} = (\mathcal{S}, \mathcal{A}, \mathcal{P}, f), \label{eq:mdp} \end{equation} where $\mathcal{S}$ and $\mathcal{A}$ are the state and action sets (\Cref{eq:state_set,eq:action_set}), $\mathcal{P}$ is the non-accessible transition function and $f$ is the reward function. The horizon is bounded by the model's maximum generation length. To solve this problem, we adopt Reinforcement Learning with Verifiable Rewards (RLVR) \citep{guo2025deepseek}, using the objective and hyperparameters of \citet{yu2026dapo} without modification. For a state $s \sim \mathcal{D}$, we sample a group of $G$ responses $\{o_i\}_{i=1}^{G} \sim \pi_{\theta_{old}}(\cdot \mid s)$, where each response $o_i$ is an action (a clue or a set of guesses) decoded as a token sequence. The policy is optimized by maximizing 
\begin{equation} 
    \label{eq:dapo-obj}
    \mathcal{J}(\theta) = \mathbb{E}_{\substack{s \sim \mathcal{D} \\ \{o_i\} \sim \pi_{\theta_{old}}}} \!\left[ \frac{1}{\sum_{i} |o_i|} \sum_{i=1}^{G} \sum_{t=1}^{|o_i|} \mathcal{L}_{i,t}(\theta) \right], 
\end{equation} 
where the per-token surrogate is 
\begin{equation} 
    \mathcal{L}_{i,t}(\theta) = \min\!\Big( r_{i,t}\, \hat{A}_{i},\; \mathrm{clip}\big(r_{i,t},\, 1\!-\!\varepsilon_{l},\, 1\!+\!\varepsilon_{h}\big)\, \hat{A}_{i} \Big) 
\end{equation} 
Here $r_{i,t} = \dfrac{\pi_\theta(o_{i,t} \mid s, o_{i,<t})}{\pi_{\theta_{old}}(o_{i,t} \mid s, o_{i,<t})}$ is the token-level importance ratio, and $\hat{A}_{i} = (R_i - \bar{R}) / \sigma_R$ is the group-normalized advantage, with $R_i = f(\cdot)$ the response reward and $\bar{R}, \sigma_R$ the mean and standard deviation of $\{R_j\}_{j=1}^{G}$. 

\section{Experimental Setup}
\label{sec:exp-setup}

\subsection{Training}
\label{sec:exp-setup-train}

\paragraph{Models.} We train three models from the Qwen3 family: Qwen3-1.7B, Qwen3-4B, and Qwen3-8B \citep{yang2025qwen3}. For every model, we perform full-parameter fine-tuning in full \texttt{bf16} precision, without parameter-efficient adapters or quantization. All training is implemented with the VeRL library \citep{sheng2025hybridflow}. For the judge model $\phi(\cdot)$, we use Google Gemini Flash-$2.5$, suitable for large inference volume. 

\paragraph{Training Algorithm and Hyperparameters.} We optimize the DAPO objective (\Cref{eq:dapo-obj}) with the GRPO advantage estimator. For each prompt, we sample a group of $G = 32$ responses, and each optimization step uses a batch of $32$ prompts, giving $32 \times 32 = 1024$ rollouts per step. We train for $2$ epochs over the $3{,}680$ training examples, yielding $230$ parameter update steps. We use a constant learning rate of $1 \times 10^{-6}$ with no warmup or decay. Following DAPO, we decouple clipping bounds with $\varepsilon_{l} = 0.2$ and $\varepsilon_{h} = 0.28$ (clip-higher), aggregate the policy loss at the token level (token-mean), and remove the KL term entirely, using neither KL loss nor KL reward shaping.

\paragraph{Dynamic Sampling and Length Control.} We adopt DAPO's dynamic sampling: we oversample and discard prompt groups whose responses all receive the same reward, since such groups yield zero advantage and no learning signal, and we allow up to $10$ oversampling batches per step to refill the effective batch. We cap the prompt length at $2,048$ tokens and the response length at $16,384$ tokens. We also apply DAPO's soft overlong reward shaping with a $4,096$-token buffer. Responses longer than $12{,}288$ tokens incur a penalty that grows linearly with the overage and reaches its maximum at the $16384$-token cap. This discourages runaway generations while avoiding hard truncation of otherwise useful reasoning traces.

\paragraph{Infrastructure.} We run experiments on a heterogeneous pool of GPUs based on availability on the Vast-AI platform. We train and evaluate Qwen3-1.7B on a single A100, train Qwen3-4B on $4{\times}$RTX 5090 and $4{\times}$RTX Pro 6000 GPUs, and train and evaluate Qwen3-8B on $4{\times}$RTX Pro 6000 GPUs. 
Approximately, the training runs for 4B and 8B models took $40$-$50$ hours on $4\times$ RTX Pro 6000 GPUs, and the 1.7B model took $\approx25$ hours on $4\times$ RTX 5090. In total, we spent approximately $3,000$ USD for training and evaluation experiments. 

\subsection{Evaluation}
\label{sec:exp-setup-eval}

\paragraph{Creativity Evaluation}
\label{sec:creativity-tasks}
We evaluate on a broad suite of creativity tasks. The suite spans divergent and convergent thinking, figurative language and humor understanding, creative problem solving, the generation of long and coherent narratives, and conceptual understanding. We group the tasks below by the construct that each one targets, and for each task, we describe its goal, a representative instance, and the metric we report.

\begin{itemize}[leftmargin=10pt, noitemsep, topsep=2pt]
    \item \emph{Divergent thinking:} \emph{AUT} (Alternate Uses Test) \citep{guilford1956structure} and \emph{DAT} (Divergent Association Test) \citep{olson2021naming} probe the ability to produce many distinct ideas. In AUT, we prompt the model to generate $5$ creative and novel uses for an everyday object, such as a belt or a brick. We score the five uses for creativity and coherence with an LLM judge, where higher is better. In DAT, we prompt the model to list $10$ words that are as semantically dissimilar from one another as possible, and we sample $100$ such lists. For each list, we compute Divergent Semantic Integration (DSI) \citep{johnson2023divergent} and report the average over all $100$ lists. We additionally report vocabulary size, the fraction of unique words across the full pool of $10 \times 100$ words. DSI captures semantic diversity, and vocabulary size captures lexical diversity; higher values are better for both.

    \item \emph{Convergent thinking:} \emph{RAT} (Remote Association Test) \citep{bowden2003normative} evaluates the complementary ability to converge on a single associative link between distant concepts. Each instance presents three remote cue words, and the model must produce one word that associates with all three. We report accuracy against the reference answers.

    \item \emph{Figurative language and humor understanding:} \emph{MUNCH} (Metaphor Understanding Challenge) \citep{tong2024metaphor} evaluates whether a model can interpret metaphorical language. It comprises three multiple-choice variants and one open-ended generation variant. We report accuracy for the multiple-choice variants and recall for the generation variant. The \emph{New Yorker Caption Contest} \citep{hessel2023androids} evaluates humor understanding. Each instance provides a description of a cartoon together with a caption. In the explanation task, the model must explain why the caption is funny, which we score with an LLM judge. In the matching task, the model selects the most suitable caption from several options, which we score with accuracy.

    \item \emph{Creative problem solving:} \emph{MacGyver} \citep{tian2024macgyver} presents an everyday problem together with a limited set of available tools, and the model must devise a creative solution that respects the tool constraints. Following \citet{tian2024macgyver}, we score solutions for feasibility and efficiency with an LLM judge. \emph{NYT Connections} \citep{lopez2025nyt} requires partitioning $16$ words into four themed groups of four. We report group-level accuracy and the exact-match rate over all four groups.

    \item \emph{Narrative generation:} \emph{CWT} (Creative Writing Test) \citep{prabhakaran2014thin} and \emph{EQ-Bench} (Emotional Quotient Benchmark) \citep{paech2023eq} target short- and long-form story writing, respectively. In CWT, we give the model three words and ask for a $5$-sentence story that uses all three, and we score the story for creativity and coherence with an LLM judge. In addition, we report DSI (semantic diversity) and MTLD (Measure of Textual Lexical Diversity) for the generated stories. In EQ-Bench, the model receives a topic and a narrative outline that specifies required details, for example \emph{``Describe the gladiator's weapon of choice and its significance to them''}, and must produce a $1{,}000$-word story. We use LLM-Judge to measure creativity and report the final score on a $0$-$100$ scale, where higher is better.

    \item \emph{Conceptual grounding:} Because grounded concept representations underpin both near and far associations, we test whether associative training strengthens them using the \emph{LLM-Grounding} dataset of \citet{xu2025large}. Each instance asks the model to rate the association between a word and a conceptual axis on an integer scale. To bound the evaluation cost, we subsample $1{,}000$ examples from the original dataset, balanced across concepts and human ratings. We compute Spearman's correlation between the model's integer ratings and the mean human ratings, where higher values indicate stronger conceptual grounding.
\end{itemize}

\input{Tables/main_results_creativity}

\paragraph{Reasoning Evaluation}
\label{sec:reasoning-tasks}
To anchor our creativity findings against well-established ability, we evaluate the same checkpoints on a standard reasoning suite covering arithmetic, competition mathematics, graduate-level science, and multi-step narrative reasoning. All tasks are run through the \texttt{lighteval} \citep{lighteval}  library with the original task formulations and metrics, and are evaluated $0$-shot. \emph{GSM8K} \citep{cobbe2021training} comprises grade-school word problems requiring multi-step arithmetic, scored by extractive-match accuracy. \emph{AIME 2024} and \emph{AIME 2025} \citep{patel2024aime} are olympiad mathematics problems with a single integer answer, also scored by extractive match. \emph{GPQA-Diamond} \citep{rein2023gpqa} is a four-way multiple-choice benchmark of Google-proof graduate-level questions in physics, chemistry, and biology, scored by accuracy. 

\paragraph{Prompts and Evaluation Hyperparameters} 
We provide exact prompts for each dataset and for all uses of LLM-Judge models (Claude Opus $4.7$) in \Cref{app:sec:evaluation-prompts}. We also provide an easy-to-read table that lists all tasks, related hyperparameters, and metrics in \Cref{app:sec:eval-data-catalog}.

\input{Tables/main_results_reasoning}

\section{Results}
\label{sec:results}
We first present our findings on the creativity tasks, and then on the reasoning benchmarks. Apart from RAT and NYT Connections, none of our evaluation tasks resemble the training task, and therefore probe the out-of-domain (OOD) transfer of our proposed setup.

\subsection{Creativity Evaluation}
\label{sec:results-creativity}
The largest model yields modest but consistent creativity improvements (\Cref{tab:finetuning-results}). Qwen3-8B improves on $20$ of $23$ creativity metrics after fine-tuning and is the only model to exhibit gains in divergent diversity, with both DAT DSI ($0.584 \to 0.589$) and vocabulary size ($0.259 \to 0.283$) increasing.

The 8B model regresses on NYT Connections despite its similarity to Codenames. Inspecting the traces reveals an overthinking tendency. With much longer reasoning, the model violates task constraints, particularly the ``\emph{one-word-one-group}'' rule, with $17$ violations for the tuned model versus $7$ for the base model. This drives the performance regression. Confirming this diagnosis, with thinking disabled, the two checkpoints (base and the finetuned version) reach an identical perfect match ($0.2730$ vs.\ $0.2730$) and group accuracy within $0.002$ ($0.4172$ vs.\ $0.4187$). On the $227$ difficulty-tagged puzzles (easy $<3$, medium $=3$, hard $\geq 4$), the tuned model degrades on every bucket with thinking on but improves on every bucket without it. The regression is, therefore, an overthinking artifact. For AUT, we find no convincing trace-level evidence explaining the minor drop in creativity.

Across scales, all three models improve on $8$ of $23$ dataset-metric pairs, and at least two improve on $15$. Gains hold at every scale for humor understanding, creative problem solving (MacGyver), story writing, and metaphor understanding (MUNCH recall). Convergent search on RAT also improves across the board, though absolute accuracy stays low for the smaller models. At the smaller scales, DAT vocabulary collapses ($1.7$B: $0.20 \to 0.10$; $4$B: $0.16 \to 0.11$), a known side effect of RLVR and RLHF training \citep{wu2025invisible, kirk2024rlhfgeneralisationdiversity}. To trace the cause, we measure average per-token entropy over the full response (thinking and final answer) alongside response length; because per-token entropy in autoregressive LMs reduces with context length, the two interact. The FT-over-base length ratios are $\times 2.45$, $\times 1.52$, and $\times 1.36$, with entropy ratios $\times 1.07$, $\times 0.69$, and $\times 1.03$ (1.7B, 4B, 8B). At 1.7B, a $7\%$ entropy gain is swamped by a $\times 2.45$ length increase. At 4B, entropy collapses ($-31\%$) while length still grows ($\times 1.52$). Only at 8B do modest entropy and length growth coincide, and DSI holds. 

\subsection{Reasoning Evaluation}
\label{sec:results-reasoning}
The reasoning results in \Cref{tab:reasoning-results} mirror the creativity picture in reverse: the gains concentrate in the smaller models. 
Qwen3-1.7B improves on all four, raising AIME24 accuracy from $0.43$ to $0.57$ and GPQA-Diamond by $5.6$ points ($0.414 \to 0.470$). Qwen3-4B improves on three, with only a marginal decrement on GSM8K ($0.9431 \to 0.9416$). The 8B model moves in the opposite direction: it holds steady on AIME25 and GSM8K but drops by roughly $2$ points on GPQA and $13$ points on AIME24 ($0.80 \to 0.67$).


\subsection{Synthesis: Associative Learning Meets RLVR}
\label{sec:results-synthesis}
Taken together, the creativity and reasoning results reveal a coherent pattern shaped by the interaction of associative learning (Codenames) with RLVR (DAPO). RLVR is known to carry a diversity-precision trade-off \citep{wu2025invisible}, and we observe exactly this, except its direction flips with scale. The 8B model shifts toward diversity, gaining creativity at a slight cost to math and science precision, while the smaller models shift the other way, gaining robustly on precision-bound reasoning without consistently gaining creativity. We speculate that associative learning sets this split \citep{johnson2025large}: easy problems are solvable by association alone, so smaller models earn reward by precisely solving the problems now within reach, while larger models, already solving these, pursue diversity for the harder ones. A length test on the 1.7B responses to the reasoning dataset supports this. Its traces lengthen overall (ratio $1.42$) but stay shorter on problems it newly solves ($1.13$) and fall below the base model on AIME ($<1.0$), converging faster on likely easy problems.
Together, these results point to a method that pays off at every scale. One associative-RLVR recipe, with no task-specific tuning, delivers substantial reasoning gains in smaller models and consistent creativity gains in larger ones. The scale-dependent trade-off is a key property that we highlight with our results: \emph{each model gains where it can}.


\section{Conclusion}
We set out to train creativity without human judgment, and found that a single game, Codenames, reshapes what a model improves at depending on its size. The same associative reward sharpens reasoning in smaller models and broadens creativity in the largest, exposing a scale-dependent precision-diversity trade-off rather than a uniform win. Concretely, fine-tuning improved the 8B model across the large majority of creativity metrics. The 1.7B and 4B models instead showed strong improvements in math and science reasoning. A single, verifiable word game thus provides a practical and scalable lever for strengthening LLMs.


\section*{Limitations}
Our work has several limitations. First, we simplify Codenames by removing the neutral and assassin words and by decomposing multi-turn play into single-turn tasks, which may omit dynamics present in the full game. Training with fewer constraints to satisfy may also explain the constraint failures we observed on NYT Connections, where the tuned model violated task rules it was never pressured to respect. Second, our training combines a chain-of-thought scaffold with generic RLVR, and, primarily due to experimental cost, we do not run an ablation that isolates the associative component; part of the observed effect may therefore stem from RLVR or the reasoning format alone. Third, we include no representational probe, which could clarify how distant concepts interact internally and how that interaction shapes the final generations. Fourth, many creativity metrics rely on an LLM judge, and the clue-generation reward itself depends on a frozen judge model, so judge bias may influence both training and evaluation. Finally, several reasoning results rest on small test sets, notably AIME with its $30$ problems, where a shift of only a few items changes accuracy substantially; these deltas should be read with corresponding caution.

\section*{Ethics Statement}
Our work focuses on training language models for creative association, improving the diversity of outputs in open-ended tasks. As our results show, creative diversity can trade off against precision, so we caution against using our models or data where factual accuracy is critical, such as medical advice, legal reasoning, or scientific fact-checking. Our training data consists of topics and word lists generated synthetically with a large language model and contains no personal, sensitive, or human-subjects data, and all evaluation benchmarks are publicly available and used in accordance with their licenses. Because several rewards and creativity metrics rely on LLM judges, which can encode cultural and linguistic biases, and because our tasks are English-centric, our results may not generalize across languages or contexts. A key motivation behind our approach is accessibility: by relying on verifiable rewards rather than human preference annotation, using a small training set, and showing gains on models as small as 1.7B, we aim to lower the resource barrier for creativity research. Finally, while we used proprietary language models (Claude Opus) to assist in experimental setup, data curation, and manuscript editing, no portion was generated entirely by an LLM; all content was written, reviewed, and edited by the authors.

\bibliography{ref_papers}

\appendix

\input{appendix}

\end{document}

%% file: appendix_prompts_preamble.tex
\usepackage{tcolorbox}
\usepackage{xcolor}
\usepackage{listings}

\tcbuselibrary{skins,breakable,listings}

\lstdefinestyle{promptstyle}{
  basicstyle=\ttfamily\footnotesize,
  breaklines=true,
  breakatwhitespace=false,
  columns=fullflexible,
  keepspaces=true,
  showstringspaces=false,
  upquote=true,
  extendedchars=true,
  literate={\`}{{\textasciigrave}}1
           {’}{{'}}1
           {‘}{{`}}1
           {“}{{"}}1
           {”}{{"}}1,
  postbreak=\mbox{\textcolor{gray}{$\hookrightarrow$}\space},
  aboveskip=0pt,
  belowskip=0pt,
}

\newtcolorbox{promptbox}[1]{%
  enhanced,
  breakable,
  colback=gray!5,
  colframe=black,
  boxrule=0.6pt,
  arc=2.5pt,
  left=6pt, right=6pt, top=4pt, bottom=4pt,
  fonttitle=\bfseries\itshape\color{white},
  coltitle=white,
  colbacktitle=black,
  attach boxed title to top left={xshift=0pt, yshift=0pt},
  boxed title style={
    colback=black,
    colframe=black,
    boxrule=0pt,
    arc=2.5pt,
    left=6pt, right=6pt, top=3pt, bottom=3pt,
  },
  title=#1,
  before upper={\par\vspace{1.2em}},
}

%% file: Tables/main_results_creativity.tex
\begin{table*}[t]
\centering
\small
\setlength{\tabcolsep}{5pt}
\renewcommand{\arraystretch}{1.15}
\begin{tabular}{@{}p{6cm} | ccc | ccc | ccc@{}}
\toprule
 & \multicolumn{3}{c}{\textbf{1.7B}} & \multicolumn{3}{c}{\textbf{4B}} & \multicolumn{3}{c}{\textbf{8B}} \\
\cmidrule(lr){2-4} \cmidrule(lr){5-7} \cmidrule(lr){8-10}
\textbf{Dataset -- Metric} & Base & Our FT & $\Delta$ & Base & Our FT & $\Delta$ & Base & Our FT & $\Delta$ \\
\midrule

AUT -- Creativity & $2.5000$ & $2.5000$ & \deq & $2.8600$ & $2.9333$ & \dup & $\mathbf{3.3733}$ & $3.3533$ & \ddn \\
AUT -- Coherence & $4.5667$ & $4.7400$ & \dup & $4.8733$ & $4.7200$ & \ddn & $4.8800$ & $\mathbf{4.9000}$ & \dup \\

\addlinespace[3pt]

DAT -- DSI & $0.5519$ & $0.5422$ & \ddn & $0.5381$ & $0.5284$ & \ddn & $0.5837$ & $\mathbf{0.5886}$ & \dup \\
DAT -- Vocab. Size & $0.2010$ & $0.1010$ & \ddn & $0.1580$ & $0.1060$ & \ddn & $0.2590$ & $\mathbf{0.2830}$ & \dup \\

\midrule

RAT -- Accuracy & $0.0554$ & $0.0590$ & \dup & $0.1144$ & $0.1255$ & \dup & $0.3432$ & $\mathbf{0.3506}$ & \dup \\

\midrule

MUNCH (Gen.) -- Recall & $0.3539$ & $0.3654$ & \dup & $0.4423$ & $0.4585$ & \dup & $0.4653$ & $\mathbf{0.4785}$ & \dup \\
MUNCH (Sent. Implicit) -- Accuracy & $0.4202$ & $0.4223$ & \dup & $0.5040$ & $0.4899$ & \ddn & $0.5282$ & $\mathbf{0.5436}$ & \dup \\
MUNCH (Sent. Mword) -- Accuracy & $0.4182$ & $0.4189$ & \dup & $0.5516$ & $0.5489$ & \ddn & $0.5724$ & $\mathbf{0.5744}$ & \dup \\
MUNCH (Word) -- Accuracy & $0.6079$ & $0.6046$ & \ddn & $0.5416$ & $\mathbf{0.5757}$ & \dup & $0.5503$ & $0.5664$ & \dup \\

\addlinespace[3pt]

NY-Caption (Explanation) -- Score & $2.0891$ & $2.1859$ & \dup & $2.4869$ & $2.5499$ & \dup & $2.7757$ & $\mathbf{2.8510}$ & \dup \\
NY-Caption (Matching) -- Accuracy & $0.4062$ & $0.4326$ & \dup & $0.5362$ & $0.5392$ & \dup & $0.5712$ & $\mathbf{0.5957}$ & \dup \\

\midrule

MacGyver -- Efficiency & $3.3548$ & $3.4921$ & \dup & $4.1064$ & $4.1642$ & \dup & $4.3557$ & $\mathbf{4.3722}$ & \dup \\
MacGyver -- Feasibility & $2.8562$ & $2.9691$ & \dup & $2.7184$ & $2.7479$ & \dup & $2.9327$ & $\mathbf{2.9412}$ & \dup \\

\addlinespace[3pt]

NYT Connections -- Group Acc. & $0.2998$ & $0.3217$ & \dup & $0.5038$ & $0.4889$ & \ddn & $\mathbf{0.6794}$ & $0.6515$ & \ddn \\
NYT Connections -- Perfect Match & $0.1150$ & $0.1166$ & \dup & $0.2991$ & $0.2761$ & \ddn & $\mathbf{0.5077}$ & $0.4724$ & \ddn \\

\midrule

CWT -- Creativity & $2.2794$ & $2.1471$ & \ddn & $2.6714$ & $2.6714$ & \deq & $2.6714$ & $\mathbf{2.8571}$ & \dup \\
CWT -- Coherence & $4.0588$ & $4.0882$ & \dup & $4.4429$ & $4.2571$ & \ddn & $4.7000$ & $\mathbf{4.7714}$ & \dup \\
CWT -- DSI & $0.7460$ & $0.7433$ & \ddn & $0.7544$ & $0.7567$ & \dup & $0.7588$ & $\mathbf{0.7612}$ & \dup \\
CWT -- MTLD & $54.4920$ & $54.0313$ & \ddn & $64.9288$ & $66.8451$ & \dup & $73.0871$ & $\mathbf{76.2662}$ & \dup \\

\addlinespace[3pt]

EQ-Bench -- Creativity & $31.4200$ & $31.7100$ & \dup & $39.2500$ & $39.6100$ & \dup & $48.2000$ & $\mathbf{48.8400}$ & \dup \\

\midrule

Grounding (Non-sensorimotor) -- Spearman-r & $0.5605$ & $0.5553$ & \ddn & $0.4827$ & $0.4568$ & \ddn & $0.4902$ & $\mathbf{0.4990}$ & \dup \\
Grounding (Sensory) -- Spearman-r & $0.4142$ & $0.4326$ & \dup & $0.5843$ & $0.5940$ & \dup & $0.6056$ & $\mathbf{0.6131}$ & \dup \\
Grounding (Motor) -- Spearman-r & $0.4196$ & $0.4104$ & \ddn & $0.5311$ & $0.5258$ & \ddn & $0.5104$ & $\mathbf{0.5345}$ & \dup \\

\bottomrule
\end{tabular}
\caption{\textbf{Creativity Evaluation.} Performance on $10$ creativity evaluation tasks across three models, Qwen3-1.7B, 4B and 8B. The ``Base'' and ``Our FT'' columns present results for off-the-shelf and trained ($230$ steps) checkpoints. The ``$\Delta$'' column indicates whether our proposed training method improves (\dup) the base model or not (\ddn).}
\label{tab:finetuning-results}
\end{table*}

%% file: Tables/main_results_reasoning.tex
\begin{table*}[t]
\centering
\small
\setlength{\tabcolsep}{5pt}
\renewcommand{\arraystretch}{1.15}
\begin{tabular}{@{}p{4cm} | ccc | ccc | ccc@{}}
\toprule
 & \multicolumn{3}{c}{\textbf{1.7B}} & \multicolumn{3}{c}{\textbf{4B}} & \multicolumn{3}{c}{\textbf{8B}} \\
\cmidrule(lr){2-4} \cmidrule(lr){5-7} \cmidrule(lr){8-10}
\textbf{Dataset -- Metric} & Base & Our FT & $\Delta$ & Base & Our FT & $\Delta$ & Base & Our FT & $\Delta$ \\
\midrule
AIME24 -- Accuracy & $0.4333$ & $0.5667$ & \dup & $0.6667$ & $0.7000$ & \dup & $0.8000$ & $0.6667$ & \ddn \\
AIME25 -- Accuracy & $0.2667$ & $0.3333$ & \dup & $0.5667$ & $0.6333$ & \dup & $0.7333$ & $0.7333$ & \deq \\
GPQA-Diamond -- Accuracy & $0.4141$ & $0.4697$ & \dup & $0.5253$ & $0.5455$ & \dup & $0.6111$ & $0.5909$ & \ddn \\
GSM8K -- Extractive Match & $0.8810$ & $0.8923$ & \dup & $0.9431$ & $0.9416$ & \ddn & $0.9507$ & $0.9500$ & \ddn \\
\bottomrule
\end{tabular}
\caption{\textbf{Reasoning Evaluation.} Performance on $4$ reasoning evaluation tasks across three models, Qwen3-1.7B, 4B and 8B. The ``Base'' and ``Our FT'' columns present results for off-the-shelf and trained ($230$ steps) checkpoints. The ``$\Delta$'' column indicates whether our proposed training method improves (\dup) the base model or not (\ddn).}
\label{tab:reasoning-results}
\end{table*}

%% file: appendix.tex
\section{Training Prompts}
\label{app:sec:prompts}

In this section, we provide the exact prompts used in our data for Codenames RLVR training. 

\begin{promptbox}{System Prompt}
\begin{lstlisting}[style=promptstyle]
You are an AI assistant that uses a structured Chain of Thought (CoT) approach to answer queries accurately and concisely.
Follow these steps in order:
1. **Think** - Identify what the problem is asking and outline your initial approach.
2. **Reason** - Work through the problem step by step, keeping each step focused and atomic.
3. **Reflect** - Check your reasoning for errors, gaps, or improvements.
4. **Adjust** - If your reflection identified an issue, explicitly correct it here. If no issues were found, state that clearly.
5. **Output** - Provide your final answer. This must be fully self-contained and readable without any context from the sections above.

---
Use the following format exactly:
---
<thinking>
[What is the problem asking? What is your initial approach?]
</thinking>

<reasoning>
[Step-by-step reasoning. Each step should be atomic and clearly follow from the previous one. Add as many steps as needed.]
</reasoning>

<reflection>
[Review your reasoning. Are there errors, missing cases, or better approaches? Be critical.]
</reflection>

<adjustment>
[If reflection found an issue: restate the corrected reasoning here.
If no issues were found: explicitly state "No adjustments needed."]
</adjustment>

<output>
[Your final, concise, self-contained answer. If uncertain, state your uncertainty clearly.]
</output>

---
Rules:
- All five sections must always appear in order.
- Only use information given in the problem. Flag missing information rather than assuming.
- Do not state conclusions before completing your reasoning.
- If considering multiple approaches, evaluate each against the original question.
- If uncertain after reflection, say so in <output> rather than guessing.
---

\end{lstlisting}
\end{promptbox}

\begin{promptbox}{User Prompt (Clue Gen.)}
\begin{lstlisting}[style=promptstyle]
Codenames Clue Generation Task: You are given a [Target-Set] and a [Non-Target-Set] of words.
Generate a single-word clue that connects to as many [Target-Set] words as possible while avoiding any association with [Non-Target-Set] words.
"Connection" is intentionally broad: semantic, conceptual, thematic, associative, phonetic, cultural, any defensible link counts.

Objective:
  - Maximize the number of [Target-Set] words your clue connects to (you may select as few as one).
  - Minimize any plausible connection between your clue and ANY [Non-Target-Set] word, under any common interpretation of the clue.
  - When these goals conflict, weigh the trade-off carefully and thoughtfully.

Clue constraints (violating any of these makes the clue invalid):
  - Must be exactly one real English word (no proper nouns, no hyphenated or compound words, no made-up words).
  - Must NOT appear in either set.
  - Must NOT be a morphological variant (inflection, derivation), abbreviation, or direct translation of any word in either set.
    Example: if SWIM is in a set, SWIMMING, SWIMMER, SWAM are all invalid.

Strategy:
  1. Brainstorm several candidate clues.
  2. For each candidate, mentally check it against every word in BOTH sets.
  3. Discard any candidate that has a plausible link to a [Non-Target-Set] word, even under uncommon meanings.
  4. Among the remaining candidates, choose the one that connects to the most [Target-Set] words.

---

Respond using exactly this format:
[CODENAMES-CLUE-START]
[Clue]: <single-word clue>
[Selected-Targets]: <list of target words your clue relates to, e.g., [WORD1, WORD2]>
[CODENAMES-CLUE-END]

---
Here are the word sets:
[Target-Set]: REFRACT, LENS
[Non-Target-Set]: TAIL, SING

\end{lstlisting}
\end{promptbox}

\begin{promptbox}{User Prompt (Guess Gen.)}
\begin{lstlisting}[style=promptstyle]
Codenames Guess Generation Task: You are given a single-word [Clue] and a list of [All-Words] on the board.
Your task is to guess which words the clue-giver intended by listing words from [All-Words] by their connection to the [Clue].
"Connection" is intentionally broad: semantic, conceptual, thematic, associative, phonetic, cultural, any defensible link counts.

Objective:
  - Identify the words in [All-Words] that the clue-giver most plausibly intended with [Clue].
  - Return up to [Max-Guesses] guesses, ordered from most confident to least confident.
  - You may return fewer than [Max-Guesses] if you are not confident in additional guesses.

Strategy:
  1. For each word in [All-Words], assess how strongly it connects to [Clue].
  2. Consider ALL possible interpretations of [Clue], the clue-giver may be using an uncommon meaning, thematic link, or lateral association.
  3. Rank candidates by connection strength. Include a word only if you believe the clue-giver plausibly chose [Clue] to point to it.
  4. Be especially cautious with lower-ranked guesses, each additional guess carries increasing risk of selecting an unintended word.

---

Respond using exactly this format:
[CODENAMES-GUESS-START]
[Guesses]: [first guess, second guess, ...]
[CODENAMES-GUESS-END]

---
Here are the word sets:
[All-Words]: BLADE, FEEL, GARDEN, LOSS
[Clue]: FROND
[Max-Guesses]: 2

\end{lstlisting}
\end{promptbox}

\section{Checks Performed on the Generated Responses}
\label{app:sec:format-reward}
Before the calculation of the reward, we first perform a comprehensive validity check on the generated responses. In \Cref{app:tab:format-checks}, we provide a list of all checks.

\input{Tables/appendix_format_reward}

\section{Evaluation Dataset Catalog}
\label{app:sec:eval-data-catalog}
In this section, we provide a catalog of dataset-specific details and hyperparameters, refer to \Cref{app:tab:eval-config}. 

\input{Tables/appendix_dataset_properties}

\section{Evaluation Prompts}
\label{app:sec:evaluation-prompts}
In this section, we provide the exact prompts used for evaluation on various creativity and reasoning datasets. 

\input{evaluation_prompts}

%% file: Tables/appendix_format_reward.tex
\begin{table}[!b]
\centering
\small
\setlength{\tabcolsep}{5pt}
\renewcommand{\arraystretch}{1.25}
\begin{tabularx}{\columnwidth}{@{}>{\raggedleft\arraybackslash}p{2.1cm} >{\raggedright\arraybackslash}X@{}}
\toprule
\textbf{Task} & \textbf{Check} \\
\midrule
\multicolumn{2}{@{}l@{}}{\textit{Clue generation}} \\
\addlinespace[2pt]
Response Format        & Clue tags (\texttt{[CODENAMES-CLUE-START], [CODENAMES-CLUE-END], [Clue], [Selected-Targets]}) are present in the generated response \\
Single Word            & Clue is a single white-space separable English word. \\
No Hyphen              & Clue does not have a hyphen \\
Non-Board Word         & Clue must not be present in the existing words ($\mathcal{B}$) on the board. \\
Morphological Variant  & Clue must not be a morphological variant of the existing words ($\mathcal{B}$) on the board. \\
Selected Targets       & Clue is generated for $1$ or more words present in the target word set $\mathcal{T}$ \\
Valid Selected Targets & Selected target words must be present in the target word set $\mathcal{T}$ \\
\addlinespace[3pt]
\midrule
\multicolumn{2}{@{}l@{}}{\textit{Guessing}} \\
\addlinespace[2pt]
Response Format & Guess tags (\texttt{[CODENAMES-GUESS-START], [CODENAMES-GUESS-END], [Guesses]}) are present in the generated response. \\
Non-empty       & There must be at least one guess generated by the model. \\
Non-Board Word  & Generated guesses must be present on the board ($\mathcal{B}$). \\
Bound           & Generated guesses must be less than the maximum number ($g_{max} = \mathcal{T}_{selected}$) provided in the input prompt. \\
\bottomrule
\end{tabularx}
\caption{Format and basic validity check on the generated responses. These checks are applied before the reward calculation.}
\label{app:tab:format-checks}
\end{table}

%% file: Tables/appendix_dataset_properties.tex
\onecolumn
\begingroup
\small
\setlength{\tabcolsep}{3pt}
\renewcommand{\arraystretch}{1.15}
\begin{longtable}{@{}
>{\raggedright\arraybackslash}p{0.12\textwidth}
>{\raggedright\arraybackslash}p{0.21\textwidth}
>{\raggedleft\arraybackslash}p{0.06\textwidth}
>{\raggedright\arraybackslash}p{0.12\textwidth}
>{\raggedright\arraybackslash}p{0.24\textwidth}
>{\raggedright\arraybackslash}p{0.17\textwidth}@{}}
\caption{Per-dataset evaluation configuration. The \emph{Generation} column lists the policy-model sampling parameters used at evaluation time (temperature $T$, nucleus $p$, top-$k$, number of samples per item $n$, and in-context examples $k$-shot). When a task uses an LLM-as-a-Judge, the judge sampling parameters are listed in italics on a second line. For all custom (creativity) tasks the policy model is run in \emph{thinking} mode (Qwen3 chat-template default) with \texttt{top\_p} and \texttt{top\_k} unset so the model's recommended thinking-mode defaults apply; for all reasoning tasks (\texttt{lighteval}) the policy is sampled in thinking mode with the default choices $T{=}0.6,\ p{=}0.95,\ n{=}1,\ k\text{-shot}{=}0,$ and a $32{,}768$-token generation budget. The judge model for all custom tasks is \texttt{Claude Opus-4.7}.}
\label{app:tab:eval-config}\\
\toprule
\textbf{Dataset} & \textbf{Goal (what it measures)} & \textbf{Size} & \textbf{Generation type} & \textbf{Generation configuration} & \textbf{Metrics} \\
\midrule
\endfirsthead
\multicolumn{6}{c}{\tablename\ \thetable\ -- continued from previous page} \\
\toprule
\textbf{Dataset} & \textbf{Goal (what it measures)} & \textbf{Size} & \textbf{Generation type} & \textbf{Generation configuration} & \textbf{Metrics} \\
\midrule
\endhead
\midrule
\multicolumn{6}{r}{\textit{Continued on next page}} \\
\endfoot
\bottomrule
\endlastfoot

\multicolumn{6}{@{}l}{\emph{Creativity tasks}} \\
\midrule
AUT & Divergent thinking: five novel uses per everyday object. & 15 & Open-ended & $T{=}1.0$, $p{=}\text{none}$, $k{=}\text{none}$, $n{=}10$, $k\text{-shot}{=}0$ \newline \textit{Judge: $T{=}0.0$, $p{=}0.95$, $n{=}1$} & LLM-judge for creativity and coherence. \\
\addlinespace[2pt]
DAT & Divergent thinking: ten maximally distant nouns. & 1 & Open-ended & $T{=}1.0$, $p{=}\text{none}$, $k{=}\text{none}$, $n{=}100$, $k\text{-shot}{=}0$ \newline \textit{(no judge)} & DSI; vocabulary size. \\
\addlinespace[2pt]
RAT & Convergent semantic search: one word linking three cues. & 271 & Open-ended (single word) & $T{=}1.0$, $p{=}\text{none}$, $k{=}\text{none}$, $n{=}1$, $k\text{-shot}{=}0$ \newline \textit{(no judge)} & Exact-match accuracy. \\
\addlinespace[2pt]
CWT & Narrative creativity: short story using three cue words. & 7 & Open-ended & $T{=}1.0$, $p{=}\text{none}$, $k{=}\text{none}$, $n{=}10$, $k\text{-shot}{=}0$ \newline \textit{Judge: $T{=}0.0$, $p{=}0.95$, $n{=}1$} & DSI; LLM-judge for creativity and coherence; Measure of Textual Lexical Diversity (MTLD). \\
\addlinespace[2pt]
EQ-Bench CW v3 & Long-form creative writing across genres. & 32 & Open-ended ($\sim$1000 words) & $T{=}1.0$, $p{=}\text{none}$, $k{=}\text{none}$, $n{=}3$, $k\text{-shot}{=}0$ \newline \textit{Judge: $T{=}0.0$, $p{=}0.95$, $n{=}1$} & 22-criterion rubric score (0--100). \\
\addlinespace[2pt]
MacGyver & Tool-use creativity: stepwise plan from a constrained tool inventory. & 1{,}683 & Open-ended & $T{=}1.0$, $p{=}\text{none}$, $k{=}\text{none}$, $n{=}1$, $k\text{-shot}{=}0$ \newline \textit{Judge: $T{=}0.0$, $p{=}0.95$, $n{=}1$} & LLM-judge feasibility / efficiency. \\
\addlinespace[2pt]
MUNCH (word) & Metaphor: 4-way MCQ word substitution. & 1{,}492 & MCQ (A/B/C/D) & $T{=}1.0$, $p{=}\text{none}$, $k{=}\text{none}$, $n{=}1$, $k\text{-shot}{=}0$ \newline \textit{(no judge)} & Accuracy (chance $25\%$). \\
\addlinespace[2pt]
MUNCH (sent-impl) & Metaphor: 4-way MCQ implicit sentence paraphrase. & 1{,}492 & MCQ (A/B/C/D) & $T{=}1.0$, $p{=}\text{none}$, $k{=}\text{none}$, $n{=}1$, $k\text{-shot}{=}0$ \newline \textit{(no judge)} & Accuracy (chance $25\%$). \\
\addlinespace[2pt]
MUNCH (sent-mword) & Metaphor: 4-way MCQ paraphrase with metaphorical-word marker. & 1{,}492 & MCQ (A/B/C/D) & $T{=}1.0$, $p{=}\text{none}$, $k{=}\text{none}$, $n{=}1$, $k\text{-shot}{=}0$ \newline \textit{(no judge)} & Accuracy (chance $25\%$). \\
\addlinespace[2pt]
MUNCH (gen) & Metaphor: open-ended one-word literal replacement. & 2{,}953 & Open-ended (single word) & $T{=}1.0$, $p{=}\text{none}$, $k{=}\text{none}$, $n{=}1$, $k\text{-shot}{=}0$ \newline \textit{(no judge)} & Recall@1 vs.\ human reference set. \\
\addlinespace[2pt]
NYC matching & Humor: 5-way MCQ identifying winning caption from description. & 2{,}691 & MCQ (A--E) & $T{=}1.0$, $p{=}\text{none}$, $k{=}\text{none}$, $n{=}1$, $k\text{-shot}{=}0$ \newline \textit{(no judge)} & Accuracy (chance $20\%$). \\
\addlinespace[2pt]
NYC ranking & Humor: 2-way MCQ funnier-of-two caption. & 2{,}616 & MCQ (A/B) & $T{=}1.0$, $p{=}\text{none}$, $k{=}\text{none}$, $n{=}1$, $k\text{-shot}{=}0$ \newline \textit{(no judge)} & Accuracy (chance $50\%$; overall, NY-editorial, crowd-rated). \\
\addlinespace[2pt]
NYC explanation & Humor: free-form explanation of a winning caption. & 651 & Open-ended (2--3 sentences) & $T{=}1.0$, $p{=}\text{none}$, $k{=}\text{none}$, $n{=}1$, $k\text{-shot}{=}0$ \newline \textit{Judge: $T{=}0.0$, $p{=}0.95$, $n{=}1$} & LLM-judge (1--5). \\
\addlinespace[2pt]
NYT Connections & Deliberate combinatorial grouping: 16 words $\rightarrow$ 4 themed groups. & 652 & Open-ended (4$\times$4 partition) & $T{=}1.0$, $p{=}\text{none}$, $k{=}\text{none}$, $n{=}1$, $k\text{-shot}{=}0$ \newline \textit{(no judge)} & Perfect-match; group accuracy. \\
\addlinespace[2pt]
LLM-Grounding & Word--concept alignment: rate words on 18 psycholinguistic dimensions. & 1{,}000 & Open-ended (integer rating) & $T{=}1.0$, $p{=}\text{none}$, $k{=}\text{none}$, $n{=}4$, $k\text{-shot}{=}0$ \newline \textit{(no judge)} & Spearman $\rho$ vs.\ human norms. \\
\midrule
\multicolumn{6}{@{}l}{\emph{Reasoning tasks (\texttt{lighteval})}} \\
\midrule
GSM8K & Grade-school multi-step arithmetic. & 1{,}319 & Open-ended & $T{=}0.6$, $p{=}0.95$, $k{=}\text{none}$, $n{=}1$, $k\text{-shot}{=}0$ & Extractive-match. \\
\addlinespace[2pt]
AIME 2024 & High-school olympiad math (integer answer). & 30 & Open-ended (boxed integer) & $T{=}0.6$, $p{=}0.95$, $k{=}\text{none}$, $n{=}1$, $k\text{-shot}{=}0$ & Extractive-match. \\
\addlinespace[2pt]
AIME 2025 & High-school olympiad math (integer answer). & 30 & Open-ended (boxed integer) & $T{=}0.6$, $p{=}0.95$, $k{=}\text{none}$, $n{=}1$, $k\text{-shot}{=}0$ & Extractive-match. \\
\addlinespace[2pt]
GPQA-Diamond & Graduate-level science MCQ (physics, chemistry, biology). & 198 & MCQ (A--D) & $T{=}0.6$, $p{=}0.95$, $k{=}\text{none}$, $n{=}1$, $k\text{-shot}{=}0$ & Accuracy (chance $25\%$). \\
\addlinespace[2pt]
MuSR murder-myst. & Multi-step suspect identification from narrative. & 250 & MCQ (2-way) & $T{=}0.6$, $p{=}0.95$, $k{=}\text{none}$, $n{=}1$, $k\text{-shot}{=}0$ & Accuracy (chance $50\%$). \\
\addlinespace[2pt]
MuSR object-plac. & Theory-of-mind: predict object search location. & 256 & MCQ (4-way) & $T{=}0.6$, $p{=}0.95$, $k{=}\text{none}$, $n{=}1$, $k\text{-shot}{=}0$ & Accuracy (chance $25\%$). \\
\addlinespace[2pt]
MuSR team-alloc. & Constraint satisfaction: assign roles to candidates. & 250 & MCQ (3-way) & $T{=}0.6$, $p{=}0.95$, $k{=}\text{none}$, $n{=}1$, $k\text{-shot}{=}0$ & Accuracy (chance ${\sim}33\%$). \\
\end{longtable}
\endgroup
\clearpage
\twocolumn

%% file: evaluation_prompts.tex
%
%
%
%


\begin{promptbox}{Alternate Uses Test (AUT)}
\begin{lstlisting}[style=promptstyle]

Prompt:

[Task Description]: For this task, you'll be asked to come up with as many original and creative uses for an object as you can.
The goal is to come up with creative ideas, which are ideas that strike people as clever, unusual, interesting, uncommon, humorous, innovative, or different.
You must provide exactly five (5) uses for the object.
---
Your final answer should be in the following format:
[Answer-Start]
[Use-1]: <less than 5 sentences describing the first use>
[Use-2]: <less than 5 sentences describing the second use>
...
[Use-5]: <less than 5 sentences describing the fifth use>
[Answer-End]
---

[Object]: belt

----------------------------

Example response

[Use-1]: A belt can be transformed into a portable plant hanger by attaching small hooks to its loops, allowing it to suspend potted plants from ceilings or doorframes.
[Use-2]: Use the belt as a makeshift drum by securing a hollow object (like a bucket) to its buckle and striking it with sticks for rhythmic beats during gatherings.
[Use-3]: Attach the belt to a skateboard or wheelchair to create a makeshift harness, enabling users to propel themselves with a push-and-pull motion.
[Use-4]: Convert the belt into a temporary fire-starting tool by rubbing the leather against a flint or steel (if dry) to generate sparks.
[Use-5]: Use the belt as a dog training tool by attaching a small leash to its buckle, allowing it to guide pets during outdoor activities.

\end{lstlisting}
\end{promptbox}

\begin{promptbox}{AUT - Judge Prompt (Creativity)}
\begin{lstlisting}[style=promptstyle]

Prompt:

You are an expert judge evaluating the CREATIVITY and DIVERSITY of proposed
uses for an everyday object.
You will be given:
[Problem]: The task instructions and the object.
[Response]: Five proposed uses for that object.

Judge ONLY creativity and diversity -- ignore grammar and writing mechanics:
  - Originality: are the uses clever, unusual, and surprising, rather than
    obvious everyday uses of the object?
  - Diversity: do the five uses span genuinely different ideas, rather than
    being minor variations on a single theme?
  - Validity: each use must be a real, physically plausible use of the
    actual object. Uses that are irrelevant to the object or impossible do
    not count as creative.

Scoring (1-5):
5 -- Five distinct, highly original uses; all valid and clearly varied.
4 -- Mostly original and varied uses, with a clear creative element.
3 -- Uses are valid but mostly conventional, or several overlap in idea.
2 -- Largely obvious uses, much repetition, or several invalid uses.
1 -- No discernible creativity, or few/no valid uses.

Provide a brief rationale (2-4 sentences), then:
[Score]: <number from 1 to 5>

[Problem]: <task instructions and object label>

[Response]: <model's five proposed uses>

----------------------------

(A second judge pass with an analogous COHERENCE rubric scores the same response on
writing mechanics. Both passes use Google gemini-2.5-flash at T=0.0, top_p=0.95, n=1.)

\end{lstlisting}
\end{promptbox}

\begin{promptbox}{Divergent Association Test (DAT)}
\begin{lstlisting}[style=promptstyle]

Prompt:

[Task description]: Please generate 10 words that are as different from each other as possible, in all meanings and uses of the words.
Rules:
- Only single words in English. Only nouns (e.g., things, objects, concepts).
- No proper nouns (e.g., no specific people or places).
- No specialized vocabulary (e.g., no technical terms).
Make a list of these 10 words, without any repetition.
---
Your final answer should be in the following format:
[Answer-Start]
[Answer]: word-1, word-2, ..., word-10
[Answer-End]
---

----------------------------

Example response

star, mountain, chair, spoon, book, love, time, freedom, smoke, fruit

\end{lstlisting}
\end{promptbox}

\begin{promptbox}{Remote Association Test (RAT)}
\begin{lstlisting}[style=promptstyle]

Prompt:

You will be given three words. Find a single word that associates with all three words.

For example: 'cottage', 'swiss', 'cake' -> 'cheese' (cottage cheese, swiss cheese, cheesecake).

Words: cottage / swiss / cake

---
Answer format: Your answer must be a single word.
[Answer-Start]
[Answer]: <solution word>
[Answer-End]
---

----------------------------

Example response

cheese

\end{lstlisting}
\end{promptbox}

\begin{promptbox}{Creative Writing Test (CWT)}
\begin{lstlisting}[style=promptstyle]

Prompt:

[Task Description]: For this task, you will write a very short story. You will be given 3 words, and must write a story that includes all 3 words. Your story should be about 5 sentences long.
Use your imagination and be creative when writing your story. But, also be sure your story makes sense.
---
Your final answer should be in the following format:
[Answer-Start]
[Answer]: <a short story, about 5 sentences long, that includes all 3 words>
[Answer-End]
---

[Words]: stamp, letter, send

----------------------------

Example response

Elena found an old stamp hidden in her grandmother's drawer, its design shimmering with forgotten magic. She wrote a letter to a distant friend, her words filled with longing. As she prepared to send it, the stamp glowed, sealing the letter in a burst of light. The letter floated through the air, carrying her message across the world. When her friend received it, they smiled, knowing it had been sent by a heart full of hope.

\end{lstlisting}
\end{promptbox}

\begin{promptbox}{CWT - Judge Prompt (Creativity)}
\begin{lstlisting}[style=promptstyle]

Prompt:

You are an expert judge evaluating the CREATIVITY of a very short story.
You will be given:
[Problem]: The task instructions and the three required words.
[Response]: A short story that should weave in all three words.

Judge ONLY creativity -- ignore grammar, spelling, and writing mechanics:
  - Originality: is the premise or idea fresh and unexpected, not cliched?
  - Imagination: are the characters, setting, or events inventive?
  - Engagement: is the story interesting, with a narrative arc or twist?
  - Use of the three words: are they woven in naturally and inventively,
    rather than dropped in mechanically? A story that ignores one or more
    of the required words cannot score above 2.

Scoring (1-5):
5 -- Highly original and imaginative; the three words are woven in cleverly.
4 -- Creative and engaging, with a clear inventive element.
3 -- Competent but conventional; a predictable idea.
2 -- Largely cliched or flat, or a required word feels forced or missing.
1 -- No discernible creativity, or required words are ignored.

Provide a brief rationale (2-4 sentences), then:
[Score]: <number from 1 to 5>

[Problem]: <task instructions and the three cue words>

[Response]: <model's short story>

----------------------------

(A second judge pass with a COHERENCE rubric scores the same response on prose
mechanics. Both passes use gemini-2.5-flash at T=0.0, top_p=0.95, n=1.)

\end{lstlisting}
\end{promptbox}

\begin{promptbox}{EQ-Bench Creative Writing v3}
\begin{lstlisting}[style=promptstyle]

Prompt:

Historical Fiction: Write a scene from a story set during the height of the Roman Empire, focusing on a slice of a day in the life of a gladiator. Do not write a combat scene. Use sensory details to capture the sights, sounds, and smells of ancient Rome. Explore the gladiator's thoughts and emotions. The story should also touch on the larger political and social issues of the time period. The piece should feel like a slice of a larger story. Describe the gladiator's weapon of choice and its significance to them. First person, past tense, 1000 words.

----------------------------

Example response (truncated; full generation ~1000 words)

**The Weight of the Spear**

The first thing I notice is the scent of sweat and iron, clinging to the air like a shroud. I'm still in my cot, the thin linen sheets damp from the night's perspiration, when the sun spills through the cracks in the arena's wooden roof. Its warmth is sharp, almost cruel, as if the gods themselves are watching me through the cracks, judging. I stretch my limbs, the stiffness in my back a familiar companion, and roll my shoulders. The sand beneath my feet is cool, gritty, and already beginning to burn where my toes press into it.

I rise, my body groaning in protest, and reach for the waterskin slung beside my cot. The water is lukewarm, flat, and smells of something I can't quite name -- like the sweat of a hundred men and the decay of the arena floor. I drink deeply, the liquid running down my throat like the last drop of a promise. My eyes flick to the spear leaning against the wall, its wooden shaft polished smooth by years of use, the iron tip gleaming dully in the light. It's my weapon, my inheritance, the only thing I have that feels like it belongs to me. My father gave it to me before he was sold into slavery, his hands trembling as he passed it down. He said it was a Thracian gift, a symbol of my people's courage. I used to believe that.

[...continues for ~700 more words spanning the training yard, the social commentary on slavery and the Roman political system, and the gladiator's reflection on whether the spear truly represents his people or only his servitude.]

\end{lstlisting}
\end{promptbox}

\begin{promptbox}{EQ-Bench - Judge Prompt}
\begin{lstlisting}[style=promptstyle]

Prompt:

You are an expert evaluator of creative writing. Read the writing prompt and the candidate story below, then judge the story's overall creativity using the four criteria below. Each criterion is defined at four quality levels (Novice = 1, Apprentice = 2, Proficient = 3, Exceptional = 4).

[Criteria]

1. Originality & Imagination
   - 4 (Exceptional): The plot features highly original, unpredictable twists. The core concept avoids cliches, offering a unique perspective or inventive treatment of the theme.
   - 3 (Proficient): Contains several imaginative ideas, though the premise or progression may follow familiar narrative tropes.
   - 2 (Apprentice): Relies heavily on familiar cliches or predictable tropes. The narrative feels derivative rather than inspired.
   - 1 (Novice): Shows little to no evidence of creative thought. The story is overly basic or entirely mimics another well-known work.

2. Sensory Detail & Word Choice (Show, Don't Tell)
   - 4 (Exceptional): Word choice is sophisticated and deliberate. The author uses vivid imagery and figurative language to immerse the reader in the story-world.
   - 3 (Proficient): Uses effective descriptions to paint a clear picture, though some moments may rely on telling rather than showing.
   - 2 (Apprentice): Lacks descriptive language. Relies on repetitive vocabulary or generalized statements rather than sensory details.
   - 1 (Novice): Little to no strength in vocabulary or description. The writing is flat and unengaging.

3. Character & Conflict Development
   - 4 (Exceptional): Characters are memorable, complex, and respond to events in highly unique ways. The conflict is compelling and drives an organic narrative arc.
   - 3 (Proficient): Main characters are adequately developed, and the primary conflict is clear and easily understood.
   - 2 (Apprentice): Characters lack depth or distinguishable personality traits. The conflict feels forced or uninteresting.
   - 1 (Novice): Characters are unidentifiable, and the central conflict is nonexistent or impossible to understand.

4. Narrative Arc & Pacing
   - 4 (Exceptional): Pacing is perfectly managed within the 1000-word constraint. Transitions between scenes are seamless, leading to an effective, satisfying resolution.
   - 3 (Proficient): Has a clear beginning, middle, and end. Ideas flow logically, though pacing may occasionally rush or drag.
   - 2 (Apprentice): Difficult to follow. Ideas or scenes feel randomly arranged, lacking clear transitions.
   - 1 (Novice): Disjointed structure. The story lacks a cohesive beginning, middle, or end.

[Writing Prompt]

{problem}

[Candidate Story]

{response}

[Instructions]

Weigh the four criteria together to form a single, comprehensive creativity judgment. Map the overall quality onto a 1-100 scale using this anchor:
  - 1-25  = Novice (all/most criteria at level 1)
  - 26-50 = Apprentice (most criteria at level 2)
  - 51-75 = Proficient (most criteria at level 3)
  - 76-100 = Exceptional (most criteria at level 4)

Be a strict but fair critic. Penalize cliched plots, telling-not-showing, flat characters, and incoherent pacing. Do not penalize NSFW or dark content; assess it on its merits.

Output format (exactly):

[Analysis]
<3-5 sentences covering the four criteria>

[Score]: <single integer between 1 and 100>

----------------------------

(The 22 criterion scores are aggregated into a Rubric Score on a 0-100 scale.
Judge is Claude Opus 4.7 at T=0.0, top_p=0.95, n=1, max_tokens=8192.)

\end{lstlisting}
\end{promptbox}

\begin{promptbox}{MacGyver}
\begin{lstlisting}[style=promptstyle]

Prompt:

[Task Description]: In this task you will be presented with a problem. Your task is to clearly understand the problem and provide a solution, if the problem is solvable. The solution you provide must be valid, feasible, efficient and must be presented in step-wise format (Step-1: ..., Step-2: ...). Use as few steps as possible and the answer should ideally be less than 100 words. If you think the problem is not solvable, provide a very short and succinct justification.
---
Final answer format should be:
[SOLUTION-START]
[Solvable]: True/False
[Solution]: (only if solvable)
Step-1: one brief sentence ...
Step-2: one brief sentence ...
...
[Justification]: (only if not solvable)
[SOLUTION-END]
---

[Problem]: You spilled red wine on the hotel carpet and want to clean it up before the housekeeping staff reports this. Tools available: a bottle opener, a plastic cup, a toothbrush, a bottle of mineral water that is sealed shut, a pack of sugar, a white bath towel, a bar of soap, a hair dryer. How do you clean up the wine stain using only these items?

----------------------------

Example response

[Solvable]: True
[Solution]: Step-1: Use the white towel to blot and absorb as much wine as possible. Step-2: Open the mineral water bottle with the opener, pour into the cup, and mix with soap to create a cleaning solution. Step-3: Scrub the stain with the toothbrush and solution. Step-4: Blot again with the towel. Step-5: Dry with the hair dryer.
[Justification]:

\end{lstlisting}
\end{promptbox}

\begin{promptbox}{MacGyver - Judge Prompt (Feasibility)}
\begin{lstlisting}[style=promptstyle]

Prompt:

You are an expert judge evaluating the FEASIBILITY of a proposed solution to a problem.
You will be given:
[Problem]: The original problem statement (states whether it is solvable).
[Response]: The model-generated answer to evaluate.

Judge ONLY feasibility -- does the proposed solution actually work given the available tools? Ignore efficiency (judged separately).

Scoring (1-5):
5 -- Correct solvability call AND a fully feasible, logically correct solution (or a correct, sufficient justification when the problem is unsolvable).
4 -- Correct solvability call with a mostly feasible solution; minor gaps or unstated assumptions.
3 -- Correct solvability call but the solution has a clear feasibility flaw (missing step, mis-used tool) that could still be patched.
2 -- Major feasibility errors: the solution would not work as written.
1 -- Mis-identifies solvability (claims a solvable problem is unsolvable or vice-versa).

Provide a brief rationale (2-4 sentences), then:
[Score]: <number from 1 to 5>

[Problem]: <problem statement, including solvability label>

[Response]: <model's stepwise solution>

----------------------------

(A second judge pass with an EFFICIENCY rubric scores the same response on path
length. Both passes use gemini-2.5-flash at T=0.0, top_p=0.95, n=1.)

\end{lstlisting}
\end{promptbox}

\begin{promptbox}{MUNCH - Word (4-way MCQ)}
\begin{lstlisting}[style=promptstyle]

Prompt:

[Task Description]: Choose which of two words best replaces a highlighted metaphorical word in a sentence without changing the meaning. You will be given a sentence and two options; respond with a single letter.

Choose the option that can replace the highlighted word in the given sentence without changing the meaning.

Sentence: While he suggests in apocalyptic *tones* in The Problem of Method that this process of self-consciousness is at last beginning to take place, and that civil, foreign and colonial wars are becoming apparent as different forms of a single class struggle, Sartre also admits that the divorce between theory and praxis which ensued under Stalinism has generally prevented any clear self-consciousness among the masses.
A. language
B. speeches
C. Both A and B
D. Neither A nor B

---
Your final answer MUST be a single letter: A, B, C, or D.
Write it wrapped in answer tags, exactly like this:
[Answer-Start]
[Answer]: <single letter A, B, C, or D>
[Answer-End]
---

----------------------------

Example response

A

\end{lstlisting}
\end{promptbox}

\begin{promptbox}{MUNCH - Judge Sentence (Implicit Paraphrase)}
\begin{lstlisting}[style=promptstyle]

Prompt:

[Task Description]: Choose which of two sentences is a paraphrase of a given sentence. You will be given the original sentence and two alternatives; respond with a single letter.

Choose the sentence that is a paraphrase of the given sentence.

Original: While he suggests in apocalyptic tones in The Problem of Method that this process of self-consciousness is at last beginning to take place, and that civil, foreign and colonial wars are becoming apparent as different forms of a single class struggle, Sartre also admits that the divorce between theory and praxis which ensued under Stalinism has generally prevented any clear self-consciousness among the masses.
A. While he suggests in apocalyptic language in The Problem of Method that this process of self-consciousness is at last beginning to take place, and that civil, foreign and colonial wars are becoming apparent as different forms of a single class struggle, Sartre also admits that the divorce between theory and praxis which ensued under Stalinism has generally prevented any clear self-consciousness among the masses.
B. While he suggests in apocalyptic speeches in The Problem of Method that this process of self-consciousness is at last beginning to take place, and that civil, foreign and colonial wars are becoming apparent as different forms of a single class struggle, Sartre also admits that the divorce between theory and praxis which ensued under Stalinism has generally prevented any clear self-consciousness among the masses.
C. Both A and B
D. Neither A nor B

---
Your final answer MUST be a single letter: A, B, C, or D.
Write it wrapped in answer tags, exactly like this:
[Answer-Start]
[Answer]: <single letter A, B, C, or D>
[Answer-End]
---

----------------------------

Example response

A

\end{lstlisting}
\end{promptbox}

\begin{promptbox}{MUNCH - Sentence (Metaphorical-Word Marker)}
\begin{lstlisting}[style=promptstyle]

Prompt:

[Task Description]: Choose which of two sentences is a paraphrase of a given sentence. You will be given the original sentence and two alternatives; respond with a single letter.

You are given a sentence in which the highlighted word is used metaphorically. Choose the sentence that is a paraphrase of it.

Original: While he suggests in apocalyptic *tones* in The Problem of Method that this process of self-consciousness is at last beginning to take place, and that civil, foreign and colonial wars are becoming apparent as different forms of a single class struggle, Sartre also admits that the divorce between theory and praxis which ensued under Stalinism has generally prevented any clear self-consciousness among the masses.
A. While he suggests in apocalyptic language in The Problem of Method that ...
B. While he suggests in apocalyptic speeches in The Problem of Method that ...
C. Both A and B
D. Neither A nor B

---
Your final answer MUST be a single letter: A, B, C, or D.
Write it wrapped in answer tags, exactly like this:
[Answer-Start]
[Answer]: <single letter A, B, C, or D>
[Answer-End]
---

----------------------------

Example response

A

\end{lstlisting}
\end{promptbox}

\begin{promptbox}{MUNCH - Generation (Single-Word Paraphrase)}
\begin{lstlisting}[style=promptstyle]

Prompt:

[Task Description]: Paraphrase a sentence by replacing a highlighted metaphorical word with a single synonym that preserves the meaning. Respond with only the replacement word.

Paraphrase the given sentence by substituting the highlighted word with another word.

Sentence: Leaders on Capitol Hill virtually put all other business on *hold* as it sought to hammer out a debt ceiling increase.

---
Your final answer MUST be a single word (the replacement for the highlighted word).
Write it wrapped in answer tags, exactly like this:
[Replacement-Word-Start]
[Replacement Word]: <single word>
[Replacement-Word-End]
---

----------------------------

Example response

pause

\end{lstlisting}
\end{promptbox}

\begin{promptbox}{New Yorker Caption Contest - Matching}
\begin{lstlisting}[style=promptstyle]

Prompt:

You are CaptionContestGPT, an expert language model at understanding the famous New Yorker caption contest. You follow the contest each week, and understand what makes for a humorous caption for each cartoon. You are aware of the various theories of humor, and read/anaylze the caption contest entries and winners each week.

Some things to remember:

- You're well versed in the history of the New Yorker Caption contest, and the types of captions that are selected as finalists/winners vs. those that are not.
- You think step-by-step, but aren't overly verbose.
- You can express uncertainty in your thinking, but in the end, pick the single best answer in the requested format.

I will describe a New Yorker cartoon to you. Then, I will give you 5 choices (labelled A-E) for captions. One of the captions was the winning caption for that cartoon, the other captions do not correspond to this cartoon. Your job is to first reason step-by-step about which answer might be correct, and, in the end, pick the single best match.

Here is a description of the cartoon followed by the five choices.

scene location: path
description: Two soldiers are on horseback in a field. A bunch of businessmen are trailing along behind them.
uncanny description: The businessmen are in the wrong time period.
entities: Mongols, Businessperson

Choices:
A: Evolution can be so tacky.
B: Let's pick up the pace. They're billing by the hour.
C: Well, five acres of popcorn back there says you were.
D: I'm sure it works. I was involved in the original research.
E: Just water for me, thanks.

Which of the 5 options (A, B, C, D, or E) is the best fit? Think step-by-step.

---
Your final answer MUST be a single letter (A, B, C, D, or E).
Write it wrapped in answer tags, exactly like this:
[Answer-Start]
[Answer]: <single letter A-E>
[Answer-End]
---

----------------------------

Example response

B

\end{lstlisting}
\end{promptbox}

\begin{promptbox}{New Yorker Caption Contest - Ranking}
\begin{lstlisting}[style=promptstyle]

Prompt:

You are CaptionContestGPT, an expert language model at understanding the famous New Yorker caption contest. ...

I will describe a New Yorker cartoon to you. Then, I will give you 2 choices (labelled A and B) for captions. One of the two captions was selected as a winning caption for that cartoon, while the other one is less funny. Your job is to first reason step-by-step about which caption might be funnier, and, in the end, select the one that you think humans found funnier.

Here is a description of the cartoon followed by the two choices.

scene location: the pyramids
description: A man is giving a hot dog to a woman outside a pyramid. On top, there is a vendor.
uncanny description: The top of a pyramid is an odd location for a vendor.
entities: Tourism, Food cart, Hot dog

Choices:
A: I'll admit, the grill marks really could pass for hieroglyphs.
B: Of course it's kosher -- who do you think built this place?

Which of the 2 options (A or B) is funnier? Think step-by-step.

---
Your final answer MUST be a single letter (A or B).
Write it wrapped in answer tags, exactly like this:
[Answer-Start]
[Answer]: <single letter A or B>
[Answer-End]
---

----------------------------

Example response

A

\end{lstlisting}
\end{promptbox}

\begin{promptbox}{New Yorker Caption Contest - Explanation}
\begin{lstlisting}[style=promptstyle]

Prompt:

You are CaptionContestGPT, an expert language model at understanding the famous New Yorker caption contest. ...

I will describe a New Yorker cartoon to you. Then, I will give you a high quality caption for that cartoon. Your job is to explain the joke in 2-3 sentences as if you were explaining to a friend who doesn't get the joke yet. First reason step-by-step about how the caption relates to the cartoon, and, in the end, write your 2-3 sentence explanation.

Here is a description of the cartoon followed by the winning caption.

scene location: mount sinai
description: A man is standing on a hill. A cloud with Groucho glasses and a hand reaching out of it is handing him a tablet.
uncanny description: The cloud has a funny disguise on.
entities: Moses, Groucho glasses, God

Winning caption: Remember, it's all in the timing.

Think step-by-step to figure out why the caption is funny for the cartoon.

---
Your final answer MUST be a 2-3 sentence explanation of the joke.
Write it wrapped in answer tags, exactly like this:
[Explanation-Start]
<your 2-3 sentence explanation>
[Explanation-End]
---

----------------------------

Example response

The cartoon reimagines Moses receiving the Ten Commandments from a cloud wearing Groucho Marx's iconic glasses, blending biblical lore with comedic absurdity. The caption "Remember, it's all in the timing" playfully nods to the classic Marx Brothers line, humorously framing the divine delivery as a timing act -- mixing religious imagery with slapstick humor and a witty pop culture reference.

\end{lstlisting}
\end{promptbox}

\begin{promptbox}{New Yorker Explanation - Judge Prompt}
\begin{lstlisting}[style=promptstyle]

Prompt:

You are an expert judge evaluating explanations of why a New Yorker caption-contest cartoon is funny.

You will be given:
[Problem]: a description of a cartoon, its winning caption, and a high-quality human-written reference explanation of the joke.
[Response]: a model-generated explanation of the same joke.

A good explanation makes the joke understandable to a friend who does not get it yet: it correctly identifies what is unexpected or incongruous in the cartoon and clearly connects the caption to that incongruity. Judge ONLY the model-generated explanation in [Response]; use the human reference only as a guide to the correct interpretation of the joke.

Score the model explanation from 1 to 5:
5 = fully correct interpretation, clearly explains the humor, as good as or better than the reference.
4 = correct interpretation, explains the humor well with minor gaps.
3 = partially correct; gets the gist but misses or muddles a key element.
2 = largely incorrect or vague; misidentifies why the joke works.
1 = wrong, irrelevant, empty, or does not explain the joke at all.

Provide a brief rationale (2-4 sentences), then state your score as:
[Score]: <number from 1 to 5>

[Problem]: <cartoon description, winning caption, and reference explanation>

[Response]: <model's 2-3 sentence explanation>

----------------------------

(Judge is Google gemini-2.5-flash at T=0.0, top_p=0.95, n=1.)

\end{lstlisting}
\end{promptbox}

\begin{promptbox}{NYT Connections}
\begin{lstlisting}[style=promptstyle]

Prompt:

[Task Description]: Group 16 words into 4 categories of 4 closely related words each. Every word belongs to exactly one group.

You are playing the New York Times Connections puzzle. In this game:
- You are given 16 words.
- Your goal is to find 4 groups of 4 words, where each group shares a specific theme or relationship.
- Every word belongs to exactly one group -- no word is left over.
- Be careful: the puzzle is designed to mislead. Some words may seem to fit multiple groups. Only one complete grouping of all 16 words is correct.
- Think carefully and deliberately before committing to a group.

---
Solve this puzzle:
Words: LASER, PLUCK, THREAD, WAX, COIL, SPOOL, WIND, WRAP, HONEYCOMB, ORGANISM, SOLAR PANEL, SPREADSHEET, BALL, MOVIE, SCHOOL, VITAMIN

Think through the possible connections. Then provide your final answer.

---
Your final answer MUST follow this exact format:
[FINAL-GROUPING-START]
[Group-1]: word1, word2, word3, word4
[Group-2]: word1, word2, word3, word4
[Group-3]: word1, word2, word3, word4
[Group-4]: word1, word2, word3, word4
[FINAL-GROUPING-END]
---

----------------------------

Example response

[FINAL-GROUPING-START]
[Group-1]: WIND, WRAP, COIL, SPOOL
[Group-2]: PLUCK, WAX, THREAD, HONEYCOMB
[Group-3]: SOLAR PANEL, SPREADSHEET, LASER, VITAMIN
[Group-4]: BALL, MOVIE, SCHOOL, ORGANISM
[FINAL-GROUPING-END]

\end{lstlisting}
\end{promptbox}

\begin{promptbox}{LLM-Grounding (Psycholinguistic Rating)}
\begin{lstlisting}[style=promptstyle]

Prompt:

[Task Description]: Rate how arousing a given word is to human beings on a numeric scale. You will be given one word and must respond with a single number.

Arousal is a measure of excitement versus calmness. A word is AROUSING if it makes you feel stimulated, excited, frenzied, jittery or wide awake. A word is UNAROUSING if it makes you feel relaxed, calm, sluggish, dull or sleepy. Please indicate how arousing human beings think each word is on a 9-point scale of VERY UNAROUSING (1) to VERY AROUSING (9), with the midpoint representing moderate arousal.

---
Word: barrage

---
End your response with your rating on its own line in this exact format:
[Rating]: <a single integer>

----------------------------

Example response

7

\end{lstlisting}
\end{promptbox}


\begin{promptbox}{GSM8K}
\begin{lstlisting}[style=promptstyle]

Prompt:

Question: Jordan has 2 children who wear diapers. Each child requires 5 diaper changes per day. Jordan's wife changes half of the diapers. How many diapers does Jordan change per day?
Answer:

----------------------------

Example response

Jordan has 2 children, each requiring 5 diaper changes per day.

**Total diaper changes per day** = 2 children x 5 changes = **10 changes**.

Jordan's wife changes half of the diapers, so Jordan changes the other half:

**Diapers Jordan changes per day** = 10 total changes / 2 = **5 changes**.

**Answer:** Jordan changes **5** diapers per day.

\end{lstlisting}
\end{promptbox}

\begin{promptbox}{AIME 2024}
\begin{lstlisting}[style=promptstyle]

Prompt:

Solve the following math problem efficiently and clearly. The last line of your response should be of the following format: 'Therefore, the final answer is: $\boxed{ANSWER}$. I hope it is correct' (without quotes) where ANSWER is just the final number or expression that solves the problem. Think step by step before answering.

Let $\omega \neq 1$ be a 13th root of unity. Find the remainder when
\[ \prod_{k=0}^{12}(2 - 2\omega^k + \omega^{2k}) \]
is divided by 1000.

----------------------------

Example response (truncated; full generation reaches the boxed integer answer)

To solve the problem, we are given a 13th root of unity $\omega \neq 1$, and we are asked to compute the remainder when the product
$$ \prod_{k=0}^{12} (2 - 2\omega^k + \omega^{2k}) $$
is divided by 1000.

### Step 1: Understanding the Structure of the Product
Let us define $P(x) = x^2 - 2x + 2$. Then the product becomes $\prod_{k=0}^{12} P(\omega^k)$, evaluated over the 13 roots of unity. Using resultant identities and the factorisation $x^{13} - 1 = \prod_k (x - \omega^k)$, the product reduces to $\text{Res}(P, x^{13}-1)$.

[...continues with computing the resultant, taking modulus 1000, and concluding:]

Therefore, the final answer is: $\boxed{321}$. I hope it is correct

\end{lstlisting}
\end{promptbox}

\begin{promptbox}{AIME 2025}
\begin{lstlisting}[style=promptstyle]

Prompt:

Solve the following math problem efficiently and clearly. The last line of your response should be of the following format: 'Therefore, the final answer is: $\boxed{ANSWER}$. I hope it is correct' (without quotes) where ANSWER is just the final number or expression that solves the problem. Think step by step before answering.

Let $N$ denote the number of ordered triples of positive integers $(a, b, c)$ such that $a, b, c \leq 3^6$ and $a^3 + b^3 + c^3$ is a multiple of $3^7$. Find the remainder when $N$ is divided by $1000$.

----------------------------

Example response (truncated)

To solve, we count ordered triples $(a, b, c)$ with $1 \leq a, b, c \leq 729$ such that $a^3 + b^3 + c^3 \equiv 0 \pmod{3^7}$. Writing each integer as $x = 3^k m$ with $\gcd(m, 3) = 1$, we have $x^3 = 3^{3k} m^3$, so the 3-adic valuation of the sum depends on the $k$-pattern of the triple.

[...case analysis over k-pattern combinations and counting residues yields the count modulo 1000.]

Therefore, the final answer is: $\boxed{735}$. I hope it is correct

\end{lstlisting}
\end{promptbox}

\begin{promptbox}{GPQA-Diamond}
\begin{lstlisting}[style=promptstyle]

Prompt:

Answer the following multiple choice question. The last line of your response should be of the following format: 'Answer: $LETTER' (without quotes) where LETTER is one of ABCD. Think step by step before answering.

In a parallel universe where a magnet can have an isolated North or South pole, Maxwell's equations look different. But, specifically, which of those equations are different?

A) The ones related to the circulation of the electric field and the divergence of the magnetic field.
B) The ones related to the divergence and the curl of the magnetic field.
C) The one related to the divergence of the magnetic field.
D) The one related to the circulation of the magnetic field and the flux of the electric field.

----------------------------

Example response

In our universe, Maxwell's equations assume the absence of magnetic monopoles, which means the divergence of the magnetic field (Gauss's Law for Magnetism) is zero: del . B = 0. In the hypothetical parallel universe where magnetic monopoles exist, this equation changes to del . B = rho_m / mu_0, where rho_m represents magnetic charge density. This modification directly alters the divergence of the magnetic field.

Other equations, such as Faraday's Law (circulation of the electric field) and Ampere's Law (curl of the magnetic field), remain unchanged unless additional terms are introduced for magnetic currents or monopole effects, which are not explicitly mentioned in the question. Therefore, the only equation that differs is the one related to the divergence of the magnetic field.

Answer: C

\end{lstlisting}
\end{promptbox}

\begin{promptbox}{MuSR - Murder Mysteries}
\begin{lstlisting}[style=promptstyle]

Prompt (truncated; full narrative ~800 words)

In a shocking twist of events, Faith has been gruesomely murdered by a lawnmower in her own kitchen and now, Detective Winston must unravel the tangled web of suspicion entangling Clyde and Paul.

[...detailed narrative describing (i) Clyde, an army veteran with weapon training, an obsessive cleanliness habit, a history of guests' items going missing, and a public-humiliation grudge against Faith; (ii) Paul, a barber with sharp-blade familiarity, regular kitchen access while doing Faith's yard work, mounting debts, and named as sole beneficiary of an unusually large insurance policy Faith recently took out...]

Who is the most likely murderer?

1 - Clyde
2 - Paul
Answer:

----------------------------

Example response (lighteval evaluates this task via log-likelihood ranking over the
provided choices; no free-form generation is produced -- the model implicitly selects
the choice with the highest conditional log-likelihood):

Model selection: 2 -> Paul (correct; means + motive + opportunity + financial windfall)

\end{lstlisting}
\end{promptbox}

\begin{promptbox}{MuSR - Object Placements}
\begin{lstlisting}[style=promptstyle]

Prompt (truncated; full narrative ~600 words)

Lisa and John were in the throes of preparation for their anticipated weekend adventure -- pedaling through nature trails on their trusted tandem bicycle.

[...narrative describes a sequence of object moves in a garage: (i) helmets start in the storage closet; (ii) Lisa carries them to a nearby bench to inspect them; (iii) Lisa leaves the garage briefly to look for sunscreen; (iv) Ellie, the housekeeper, returns the helmets from the bench back into the storage closet while Lisa is away; (v) Lisa returns. The question asks where Lisa will look for the helmets, requiring tracking what Lisa observed vs. what Ellie did out of her sight...]

Which location is the most likely place Lisa would look to find the helmets given the story?

1 - bench
2 - driveway
3 - storage closet
4 - wall
Answer:

----------------------------

Example response (log-likelihood selection over choices):

Model selection: 3 -> storage closet (correct)

\end{lstlisting}
\end{promptbox}

\begin{promptbox}{MuSR - Team Allocation}
\begin{lstlisting}[style=promptstyle]

Prompt (truncated; full narrative ~600 words)

Welcome to the story of a charming, fledgling diner, where the scent of new upholstery mingled with the promise of fresh beginnings.

[...character profiles: Jessica is a culinary perfectionist with award-winning cooking skill but clashes with the temperaments of both co-workers; Rebecca is a calm cook who is also exceptional with customer memory and service; Samuel is a former kitchen assistant who waited tables at a five-star restaurant and works best under Rebecca's mentorship. The puzzle requires choosing the allocation of chef vs. server roles that best accounts for skill fit and interpersonal dynamics...]

Given the story, how would you uniquely allocate each person to make sure both tasks are accomplished efficiently?

1 - Cooking meals: Samuel, Serving customers: Jessica and Rebecca
2 - Cooking meals: Rebecca, Serving customers: Jessica and Samuel
3 - Cooking meals: Jessica, Serving customers: Rebecca and Samuel
Answer:

----------------------------

Example response (log-likelihood selection over choices):

Model selection: 3 -> Cooking meals: Jessica, Serving customers: Rebecca and Samuel (correct)

\end{lstlisting}
\end{promptbox}

%% file: ref_papers.bib
@article{bubeck2023paper,
  title={Paper Review:'Sparks of Artificial General Intelligence: Early experiments with GPT-4'},
  author={Bubeck, S{\'e}bastien and Chandrasekaran, Varun and Eldan, Ronen and Gehrke, Johannes and Horvitz, Eric and Kamar, Ece and Lee, Peter and Lee, Yin Tat and Li, Yuanzhi and Lundberg, Scott and others},
  year={2023}
}

@inproceedings{ahn2024large,
  title={Large language models for mathematical reasoning: Progresses and challenges},
  author={Ahn, Janice and Verma, Rishu and Lou, Renze and Liu, Di and Zhang, Rui and Yin, Wenpeng},
  booktitle={Proceedings of the 18th Conference of the European Chapter of the Association for Computational Linguistics: Student Research Workshop},
  pages={225--237},
  year={2024}
}

@article{wu2026writingbench,
  title={Writingbench: A comprehensive benchmark for generative writing},
  author={Wu, Yuning and Mei, Jiahao and Yan, Ming and Li, Chenliang and Lai, Shaopeng and Ren, Yuran and Wang, Zijia and Zhang, Ji and Wu, Mengyue and Jin, Qin and others},
  journal={Advances in Neural Information Processing Systems},
  volume={38},
  year={2026}
}

@article{mednick1962associative,
  title={The associative basis of the creative process.},
  author={Mednick, Sarnoff},
  journal={Psychological review},
  volume={69},
  number={3},
  pages={220},
  year={1962},
  publisher={American Psychological Association}
}

@article{beaty2023associative,
  title={Associative thinking at the core of creativity},
  author={Beaty, Roger E and Kenett, Yoed N},
  journal={Trends in cognitive sciences},
  volume={27},
  number={7},
  pages={671--683},
  year={2023},
  publisher={Elsevier}
}

@article{luchini2025creativity,
  title={Creativity supports learning through associative thinking},
  author={Luchini, Simone A and Kaufman, James C and Goecke, Benjamin and Wilhelm, Oliver and Kenett, Yoed N and Lei, Daisy and Benedek, Mathias and van Hell, Janet G and Beaty, Roger E},
  journal={npj Science of Learning},
  volume={10},
  number={1},
  pages={42},
  year={2025},
  publisher={Nature Publishing Group UK London}
}

@article{kenett2014investigating,
  title={Investigating the structure of semantic networks in low and high creative persons},
  author={Kenett, Yoed N and Anaki, David and Faust, Miriam},
  journal={Frontiers in human neuroscience},
  volume={8},
  pages={407},
  year={2014},
  publisher={Frontiers Media SA}
}

@article{kenett2019semantic,
  title={A semantic network cartography of the creative mind},
  author={Kenett, Yoed N and Faust, Miriam},
  journal={Trends in cognitive sciences},
  volume={23},
  number={4},
  pages={271--274},
  year={2019},
  publisher={Elsevier}
}

@article{boden1998creativity,
  title={Creativity and artificial intelligence},
  author={Boden, Margaret A},
  journal={Artificial intelligence},
  volume={103},
  number={1-2},
  pages={347--356},
  year={1998},
  publisher={Elsevier}
}

@article{yang2022creative,
  title={Creative problem solving in knowledge-rich contexts},
  author={Yang, Wenjing and Green, Adam E and Chen, Qunlin and Kenett, Yoed N and Sun, Jiangzhou and Wei, Dongtao and Qiu, Jiang},
  journal={Trends in Cognitive Sciences},
  volume={26},
  number={10},
  pages={849--859},
  year={2022},
  publisher={Elsevier}
}

@article{mikolov2013distributed,
  title={Distributed representations of words and phrases and their compositionality},
  author={Mikolov, Tomas and Sutskever, Ilya and Chen, Kai and Corrado, Greg S and Dean, Jeff},
  journal={Advances in neural information processing systems},
  volume={26},
  year={2013}
}

@inproceedings{johnson2025large,
  title={Do large language models solve verbal analogies like children do?},
  author={Johnson, Tamar and ter Veen, Mathilde and Choenni, Rochelle and van der Maas, Han and Shutova, Ekaterina and Stevenson, Claire E},
  booktitle={Proceedings of the 29th Conference on Computational Natural Language Learning},
  pages={627--639},
  year={2025}
}

@article{barenholtz2026world,
  title={World Properties without World Models: Recovering Spatial and Temporal Structure from Co-occurrence Statistics in Static Word Embeddings},
  author={Barenholtz, Elan},
  journal={arXiv preprint arXiv:2603.04317},
  year={2026}
}

@article{vessel2018stronger,
  title={Stronger shared taste for natural aesthetic domains than for artifacts of human culture},
  author={Vessel, Edward A and Maurer, Natalia and Denker, Alexander H and Starr, G Gabrielle},
  journal={Cognition},
  volume={179},
  pages={121--131},
  year={2018},
  publisher={Elsevier}
}

@article{forthmann2017missing,
  title={Missing creativity: The effect of cognitive workload on rater (dis-) agreement in subjective divergent-thinking scores},
  author={Forthmann, Boris and Holling, Heinz and Zandi, Nima and Gerwig, Anne and {\c{C}}elik, P{\i}nar and Storme, Martin and Lubart, Todd},
  journal={Thinking Skills and Creativity},
  volume={23},
  pages={129--139},
  year={2017},
  publisher={Elsevier}
}

@article{cortes2021inconsistency,
  title={Inconsistency in conference peer review: Revisiting the 2014 neurips experiment},
  author={Cortes, Corinna and Lawrence, Neil D},
  journal={arXiv preprint arXiv:2109.09774},
  year={2021}
}

@article{casper2023open,
  title={Open problems and fundamental limitations of reinforcement learning from human feedback},
  author={Casper, Stephen and Davies, Xander and Shi, Claudia and Gilbert, Thomas Krendl and Scheurer, J{\'e}r{\'e}my and Rando, Javier and Freedman, Rachel and Korbak, Tomasz and Lindner, David and Freire, Pedro and others},
  journal={arXiv preprint arXiv:2307.15217},
  year={2023}
}

@article{cropley2006praise,
  title={In praise of convergent thinking},
  author={Cropley, Arthur},
  journal={Creativity research journal},
  volume={18},
  number={3},
  pages={391--404},
  year={2006},
  publisher={Taylor \& Francis}
}

@article{guo2025deepseek,
  title={Deepseek-r1: Incentivizing reasoning capability in llms via reinforcement learning},
  author={Guo, Daya and Yang, Dejian and Zhang, Haowei and Song, Junxiao and Wang, Peiyi and Zhu, Qihao and Xu, Runxin and Zhang, Ruoyu and Ma, Shirong and Bi, Xiao and others},
  journal={arXiv preprint arXiv:2501.12948},
  year={2025}
}

@article{yu2026dapo,
  title={Dapo: An open-source llm reinforcement learning system at scale},
  author={Yu, Qiying and Zhang, Zheng and Zhu, Ruofei and Yuan, Yufeng and Zuo, Xiaochen and Yue, Yu and Dai, Weinan and Fan, Tiantian and Liu, Gaohong and Liu, Lingjun and others},
  journal={Advances in Neural Information Processing Systems},
  volume={38},
  pages={113222--113244},
  year={2026}
}

@inproceedings{sheng2025hybridflow,
  title={Hybridflow: A flexible and efficient rlhf framework},
  author={Sheng, Guangming and Zhang, Chi and Ye, Zilingfeng and Wu, Xibin and Zhang, Wang and Zhang, Ru and Peng, Yanghua and Lin, Haibin and Wu, Chuan},
  booktitle={Proceedings of the Twentieth European Conference on Computer Systems},
  pages={1279--1297},
  year={2025}
}

@article{liu2025spiral,
  title={Spiral: Self-play on zero-sum games incentivizes reasoning via multi-agent multi-turn reinforcement learning},
  author={Liu, Bo and Guertler, Leon and Yu, Simon and Liu, Zichen and Qi, Penghui and Balcells, Daniel and Liu, Mickel and Tan, Cheston and Shi, Weiyan and Lin, Min and others},
  journal={arXiv preprint arXiv:2506.24119},
  year={2025}
}

@article{cheng2024self,
  title={Self-playing adversarial language game enhances llm reasoning},
  author={Cheng, Pengyu and Dai, Yong and Hu, Tianhao and Xu, Han and Zhang, Zhisong and Han, Lei and Du, Nan and Li, Xiaolong},
  journal={Advances in Neural Information Processing Systems},
  volume={37},
  pages={126515--126543},
  year={2024}
}

@article{wang2025vision,
  title={Vision-zero: Scalable vlm self-improvement via strategic gamified self-play},
  author={Wang, Qinsi and Liu, Bo and Zhou, Tianyi and Shi, Jing and Lin, Yueqian and Chen, Yiran and Li, Hai Helen and Wan, Kun and Zhao, Wentian},
  journal={arXiv preprint arXiv:2509.25541},
  year={2025}
}

@article{xie2025logic,
  title={Logic-rl: Unleashing llm reasoning with rule-based reinforcement learning},
  author={Xie, Tian and Gao, Zitian and Ren, Qingnan and Luo, Haoming and Hong, Yuqian and Dai, Bryan and Zhou, Joey and Qiu, Kai and Wu, Zhirong and Luo, Chong},
  journal={arXiv preprint arXiv:2502.14768},
  year={2025}
}

@article{schaul2024boundless,
  title={Boundless socratic learning with language games},
  author={Schaul, Tom},
  journal={arXiv preprint arXiv:2411.16905},
  year={2024}
}

@article{wong2025logicpuzzlerl,
  title={LogicPuzzleRL: cultivating robust mathematical reasoning in LLMs via reinforcement learning},
  author={Wong, Zhen Hao and Deng, Jingwen and He, Runming and Chen, Zirong and You, Qijie and Dong, Hejun and Liang, Hao and Shen, Chengyu and Cui, Bin and Zhang, Wentao},
  journal={arXiv preprint arXiv:2506.04821},
  year={2025}
}

@article{hwang2025can,
  title={Can Large Language Models Develop Strategic Reasoning? Post-training Insights from Learning Chess},
  author={Hwang, Dongyoon and Lee, Hojoon and Choo, Jaegul and Park, Dongmin and Park, Jongho},
  journal={arXiv preprint arXiv:2507.00726},
  year={2025}
}

@article{yuan2025marshal,
  title={Marshal: Incentivizing multi-agent reasoning via self-play with strategic llms},
  author={Yuan, Huining and Xu, Zelai and Tan, Zheyue and Yi, Xiangmin and Guang, Mo and Long, Kaiwen and Hui, Haojia and Li, Boxun and Chen, Xinlei and Zhao, Bo and others},
  journal={arXiv preprint arXiv:2510.15414},
  year={2025}
}

@article{xie2025play,
  title={Play to generalize: Learning to reason through game play},
  author={Xie, Yunfei and Ma, Yinsong and Lan, Shiyi and Yuille, Alan and Xiao, Junfei and Wei, Chen},
  journal={arXiv preprint arXiv:2506.08011},
  year={2025}
}

@article{tong2025game,
  title={Game-RL: Synthesizing Multimodal Verifiable Game Data to Boost VLMs' General Reasoning},
  author={Tong, Jingqi and Tang, Jixin and Li, Hangcheng and Mou, Yurong and Zhang, Ming and Zhao, Jun and Wen, Yanbo and Song, Fan and Zhan, Jiahao and Lu, Yuyang and others},
  journal={arXiv preprint arXiv:2505.13886},
  year={2025}
}

@article{guilford1956structure,
  title={The structure of intellect.},
  author={Guilford, Joy Paul},
  journal={Psychological bulletin},
  volume={53},
  number={4},
  pages={267},
  year={1956},
  publisher={American Psychological Association}
}

@article{olson2021naming,
  title={Naming unrelated words predicts creativity},
  author={Olson, Jay A and Nahas, Johnny and Chmoulevitch, Denis and Cropper, Simon J and Webb, Margaret E},
  journal={Proceedings of the National Academy of Sciences},
  volume={118},
  number={25},
  pages={e2022340118},
  year={2021},
  publisher={National Academy of Sciences}
}

@article{prabhakaran2014thin,
  title={Thin slices of creativity: Using single-word utterances to assess creative cognition},
  author={Prabhakaran, Ranjani and Green, Adam E and Gray, Jeremy R},
  journal={Behavior research methods},
  volume={46},
  number={3},
  pages={641--659},
  year={2014},
  publisher={Springer}
}

@article{bowden2003normative,
  title={Normative data for 144 compound remote associate problems},
  author={Bowden, Edward M and Jung-Beeman, Mark},
  journal={Behavior research methods, instruments, \& computers},
  volume={35},
  number={4},
  pages={634--639},
  year={2003},
  publisher={Springer}
}

@article{paech2023eq,
  title={Eq-bench: An emotional intelligence benchmark for large language models},
  author={Paech, Samuel J},
  journal={arXiv preprint arXiv:2312.06281},
  year={2023}
}

@inproceedings{tian2024macgyver,
  title={MacGyver: Are Large Language Models Creative Problem Solvers?},
  author={Tian, Yufei and Ravichander, Abhilasha and Qin, Lianhui and Le Bras, Ronan and Marjieh, Raja and Peng, Nanyun and Choi, Yejin and Griffiths, Thomas L and Brahman, Faeze},
  booktitle={Proceedings of the 2024 Conference of the North American Chapter of the Association for Computational Linguistics: Human Language Technologies (Volume 1: Long Papers)},
  pages={5303--5324},
  year={2024}
}

@inproceedings{tong2024metaphor,
  title={Metaphor understanding challenge dataset for LLMs},
  author={Tong, Xiaoyu and Choenni, Rochelle and Lewis, Martha and Shutova, Ekaterina},
  booktitle={Proceedings of the 62nd Annual Meeting of the Association for Computational Linguistics (Volume 1: Long Papers)},
  pages={3517--3536},
  year={2024}
}

@inproceedings{hessel2023androids,
  title={Do androids laugh at electric sheep? humor “understanding” benchmarks from the new yorker caption contest},
  author={Hessel, Jack and Marasovi{\'c}, Ana and Hwang, Jena D and Lee, Lillian and Da, Jeff and Zellers, Rowan and Mankoff, Robert and Choi, Yejin},
  booktitle={Proceedings of the 61st Annual Meeting of the Association for Computational Linguistics (Volume 1: Long Papers)},
  pages={688--714},
  year={2023}
}

@inproceedings{lopez2025nyt,
  title={NYT-connections: A deceptively simple text classification task that stumps system-1 thinkers},
  author={Lopez, Angel Yahir Loredo and McDonald, Tyler and Emami, Ali},
  booktitle={Proceedings of the 31st International Conference on Computational Linguistics},
  pages={1952--1963},
  year={2025}
}

@article{xu2025large,
  title={Large language models without grounding recover non-sensorimotor but not sensorimotor features of human concepts},
  author={Xu, Qihui and Peng, Yingying and Nastase, Samuel A and Chodorow, Martin and Wu, Minghua and Li, Ping},
  journal={Nature human behaviour},
  volume={9},
  number={9},
  pages={1871--1886},
  year={2025},
  publisher={Nature Publishing Group UK London}
}

@article{johnson2023divergent,
  title={Divergent semantic integration (DSI): Extracting creativity from narratives with distributional semantic modeling},
  author={Johnson, Dan R and Kaufman, James C and Baker, Brendan S and Patterson, John D and Barbot, Baptiste and Green, Adam E and van Hell, Janet and Kennedy, Evan and Sullivan, Grace F and Taylor, Christa L and others},
  journal={Behavior research methods},
  volume={55},
  number={7},
  pages={3726--3759},
  year={2023},
  publisher={Springer}
}

@article{cobbe2021training,
  title={Training verifiers to solve math word problems},
  author={Cobbe, Karl and Kosaraju, Vineet and Bavarian, Mohammad and Chen, Mark and Jun, Heewoo and Kaiser, Lukasz and Plappert, Matthias and Tworek, Jerry and Hilton, Jacob and Nakano, Reiichiro and others},
  journal={arXiv preprint arXiv:2110.14168},
  year={2021}
}

@article{patel2024aime,
  title={Aime: Ai system optimization via multiple llm evaluators},
  author={Patel, Bhrij and Chakraborty, Souradip and Suttle, Wesley A and Wang, Mengdi and Bedi, Amrit Singh and Manocha, Dinesh},
  journal={arXiv preprint arXiv:2410.03131},
  year={2024}
}

@article{rein2023gpqa,
  title={Gpqa: A graduate-level google-proof q\&a benchmark},
  author={Rein, David and Hou, Betty Li and Stickland, Asa Cooper and Petty, Jackson and Pang, Richard Yuanzhe and Dirani, Julien and Michael, Julian and Bowman, Samuel R},
  journal={arXiv preprint arXiv:2311.12022},
  year={2023}
}

@article{yang2025qwen3,
  title={Qwen3 technical report},
  author={Yang, An and Li, Anfeng and Yang, Baosong and Zhang, Beichen and Hui, Binyuan and Zheng, Bo and Yu, Bowen and Gao, Chang and Huang, Chengen and Lv, Chenxu and others},
  journal={arXiv preprint arXiv:2505.09388},
  year={2025}
}

@article{lu2024llm,
  title={Llm discussion: Enhancing the creativity of large language models via discussion framework and role-play},
  author={Lu, Li-Chun and Chen, Shou-Jen and Pai, Tsung-Min and Yu, Chan-Hung and Lee, Hung-yi and Sun, Shao-Hua},
  journal={arXiv preprint arXiv:2405.06373},
  year={2024}
}

@inproceedings{fukuda-etal-2025-building,
    title = "Building {J}apanese Creativity Benchmarks and Applying them to Enhance {LLM} Creativity",
    author = "Fukuda, So  and
      Ogawa, Hayato  and
      Horio, Kaito  and
      Kawahara, Daisuke  and
      Shibata, Tomohide",
    editor = "Zhao, Jin  and
      Wang, Mingyang  and
      Liu, Zhu",
    booktitle = "Proceedings of the 63rd Annual Meeting of the Association for Computational Linguistics (Volume 4: Student Research Workshop)",
    month = jul,
    year = "2025",
    address = "Vienna, Austria",
    publisher = "Association for Computational Linguistics",
    url = "https://aclanthology.org/2025.acl-srw.69/",
    doi = "10.18653/v1/2025.acl-srw.69",
    pages = "939--957",
    ISBN = "979-8-89176-254-1"
}

@article{shen2026unlocking,
  title={Unlocking LLM Creativity in Science through Analogical Reasoning},
  author={Shen, Andrew and Druckmann, Shaul and Zou, James},
  journal={arXiv preprint arXiv:2605.11258},
  year={2026}
}

@article{ismayilzada2025creative,
  title={Creative preference optimization},
  author={Ismayilzada, Mete and Laverghetta Jr, Antonio and Luchini, Simone A and Patel, Reet and Bosselut, Antoine and Van Der Plas, Lonneke and Beaty, Roger},
  journal={arXiv preprint arXiv:2505.14442},
  year={2025}
}

@article{mehrotra2024enhancing,
  title={Enhancing creativity in large language models through associative thinking strategies},
  author={Mehrotra, Pronita and Parab, Aishni and Gulwani, Sumit},
  journal={arXiv preprint arXiv:2405.06715},
  year={2024}
}

@inproceedings{nair-etal-2024-creative,
    title = "Creative Problem Solving in Large Language and Vision Models - What Would it Take?",
    author = "Nair, Lakshmi  and
      Gizzi, Evana  and
      Sinapov, Jivko",
    editor = "Al-Onaizan, Yaser  and
      Bansal, Mohit  and
      Chen, Yun-Nung",
    booktitle = "Findings of the Association for Computational Linguistics: EMNLP 2024",
    month = nov,
    year = "2024",
    address = "Miami, Florida, USA",
    publisher = "Association for Computational Linguistics",
    url = "https://aclanthology.org/2024.findings-emnlp.700/",
    doi = "10.18653/v1/2024.findings-emnlp.700",
    pages = "11978--11994"
}

@inproceedings{
summers-stay2023brainstorm,
title={Brainstorm, then Select: a Generative Language Model Improves Its Creativity Score},
author={Douglas Summers-Stay and Clare R. Voss and Stephanie M. Lukin},
booktitle={The AAAI-23 Workshop on Creative AI Across Modalities},
year={2023},
url={https://openreview.net/forum?id=8HwKaJ1wvl}
}

@article{franceschelli2024creative,
  title={Creative beam search: LLM-as-a-Judge for improving response generation},
  author={Franceschelli, Giorgio and Musolesi, Mirco},
  journal={arXiv preprint arXiv:2405.00099},
  year={2024}
}

@article{meister2023locally,
  title={Locally typical sampling},
  author={Meister, Clara and Pimentel, Tiago and Wiher, Gian and Cotterell, Ryan},
  journal={Transactions of the Association for Computational Linguistics},
  volume={11},
  pages={102--121},
  year={2023}
}

@misc{lighteval,
  author = {Habib, Nathan and Fourrier, Clémentine and Kydlíček, Hynek and Wolf, Thomas and Tunstall, Lewis},
  title = {LightEval: A lightweight framework for LLM evaluation},
  year = {2023},
  version = {0.11.0},
  url = {https://github.com/huggingface/lighteval}
}

@inproceedings{kirk2024rlhfgeneralisationdiversity,
  title={Understanding the effects of rlhf on llm generalisation and diversity},
  author={Kirk, Robert and Mediratta, Ishita and Nalmpantis, Christoforos and Luketina, Jelena and Hambro, Eric and Grefenstette, Edward and Raileanu, Roberta},
  booktitle={International Conference on Learning Representations},
  volume={2024},
  pages={20620--20653},
  year={2024}
}

@article{wu2025invisible,
  title={The invisible leash: Why rlvr may or may not escape its origin},
  author={Wu, Fang and Xuan, Weihao and Lu, Ximing and Liu, Mingjie and Dong, Yi and Harchaoui, Zaid and Choi, Yejin},
  journal={arXiv preprint arXiv:2507.14843},
  year={2025}
}

@article{lanchantin2025diverse,
  title={Diverse preference optimization},
  author={Lanchantin, Jack and Chen, Angelica and Dhuliawala, Shehzaad and Yu, Ping and Weston, Jason and Sukhbaatar, Sainbayar and Kulikov, Ilia},
  journal={arXiv preprint arXiv:2501.18101},
  year={2025}
}

@inproceedings{deshpande2025diverse,
  title={Diverse, not Short: A Length-Controlled Data Selection Strategy for Improving Response Diversity of Language Models},
  author={Deshpande, Vijeta and Ghose, Debasmita and Patterson, John D and Beaty, Roger E and Rumshisky, Anna},
  booktitle={Proceedings of the 2025 Conference on Empirical Methods in Natural Language Processing},
  pages={33905--33926},
  year={2025}
}

@article{ouyang2022training,
  title={Training language models to follow instructions with human feedback},
  author={Ouyang, Long and Wu, Jeffrey and Jiang, Xu and Almeida, Diogo and Wainwright, Carroll and Mishkin, Pamela and Zhang, Chong and Agarwal, Sandhini and Slama, Katarina and Ray, Alex and others},
  journal={Advances in neural information processing systems},
  volume={35},
  pages={27730--27744},
  year={2022}
}
